%% file: main.tex
\documentclass[]{bytedance_seed}

\usepackage[toc,page,header]{appendix}

\usepackage{minitoc}
\usepackage{solarized-light}
\usepackage{multirow}
\usepackage{graphicx}
\usepackage{array}
\usepackage{makecell}
\usepackage{framed}
\usepackage{hyperref}
\usepackage{amsmath} 
\usepackage[colorinlistoftodos]{todonotes}
\usepackage{longtable}
\usepackage{hhline}
\usepackage{fancyvrb}
\usepackage{fvextra}
\usepackage{CJKutf8}
\usepackage{multicol}
\usepackage{cleveref}
\usepackage{tablefootnote}
\usepackage{threeparttable}
\usepackage{tabularx}
\usepackage{mdframed}
\usepackage{subcaption}
\usepackage[usestackEOL]{stackengine}
\usepackage[authoryear]{natbib}
\newcommand{\commentout}[1]{}
\usepackage{float}       
\usepackage{afterpage}   
\renewcommand{\paragraph}[1]{\noindent\textbf{#1.}\hspace*{1em}}
\usepackage{enumitem}
\setlist[itemize]{leftmargin=15pt}
\usepackage{amsmath,amssymb,amsfonts}
\usepackage{booktabs}
\usepackage{array}
\usepackage{xspace}
\usepackage{tabu}
\usepackage{enumitem}
\usepackage{subcaption}
\usepackage[most]{tcolorbox}
\usepackage{listings}
\usepackage{xcolor}
\usepackage{cuted}
\usepackage{colortbl}      

\usepackage[absolute,overlay]{textpos}
\RequirePackage{xspace}
\makeatletter
\DeclareRobustCommand\onedot{\futurelet\@let@token\@onedot}
\def\@onedot{\ifx\@let@token.\else.\null\fi\xspace}

\makeatother

\newcommand{\grayrow}[1]{\cellcolor{gray!10}#1}

\title{PASK: \space Toward \space Intent-Aware \space Proactive \space Agents \space  \\ with \space Long-Term Memory}

\author[2 \dagger]{Zhifei Xie}
\author[1]{Zongzheng Hu}
\author[3]{Fangda Ye}
\author[1]{Xin Zhang}
\author[1]{Haobo Chai}
\author[3]{Zihang Liu}
\author[2]{Pengcheng Wu}
\author[3]{Guibin Zhang}
\author[3]{Yue Liao}
\author[3 \ddagger]{Xiaobin~Hu} 
\author[2 \ddagger]{Deheng Ye}
\author[2 \ddagger]{Chunyan Miao}
\author[3 \ddagger]{Shuicheng Yan}

\affiliation[1]{Pask-Core}
\affiliation[2]{NTU}
\affiliation[3]{NUS}

\contribution[\dagger]{Project leader}
\contribution[\ddagger]{Corresponding authors}

\abstract{
Proactivity is a core expectation for AGI. Prior work remains largely confined to laboratory settings, leaving a clear gap in real-world proactive agent: depth, complexity, ambiguity, precision and real-time constraints. We study this setting, where useful intervention requires inferring latent needs from ongoing context and grounding actions in evolving user memory under latency and long-horizon constraints. We first propose DD-MM-PAS (Demand Detection, Memory Modeling, Proactive Agent System) as a general paradigm for streaming proactive ai agent. We instantiate this paradigm in Pask, with streaming IntentFlow model for DD, a hybrid memory (workspace, user, global) for long-term MM, PAS infra framework and introduce how these components form a closed loop. We also introduce LatentNeeds-Bench, a real-world benchmark built from user-consented data and refined through thousands of rounds of human editing. Experiments show that IntentFlow matches leading Gemini3-Flash models under latency constraints, while identifying deeper user intent.
}

\correspondence{\href{https://xzf-thu.github.io/Pask/}
{\textbf{https://xzf-thu.github.io/Pask/}}}

\correspondencee{\texttt{zhifei001@e.ntu.edu.sg} \quad \texttt{ascymiao@ntu.edu.sg} \quad \texttt{yansc@nus.edu.sg}}

\begin{document}

\maketitle

\begin{textblock*}{180mm}(18.1mm,170mm)
   \label{fig:1-paradigm comparison}
    \centering
    \includegraphics[width=175mm]{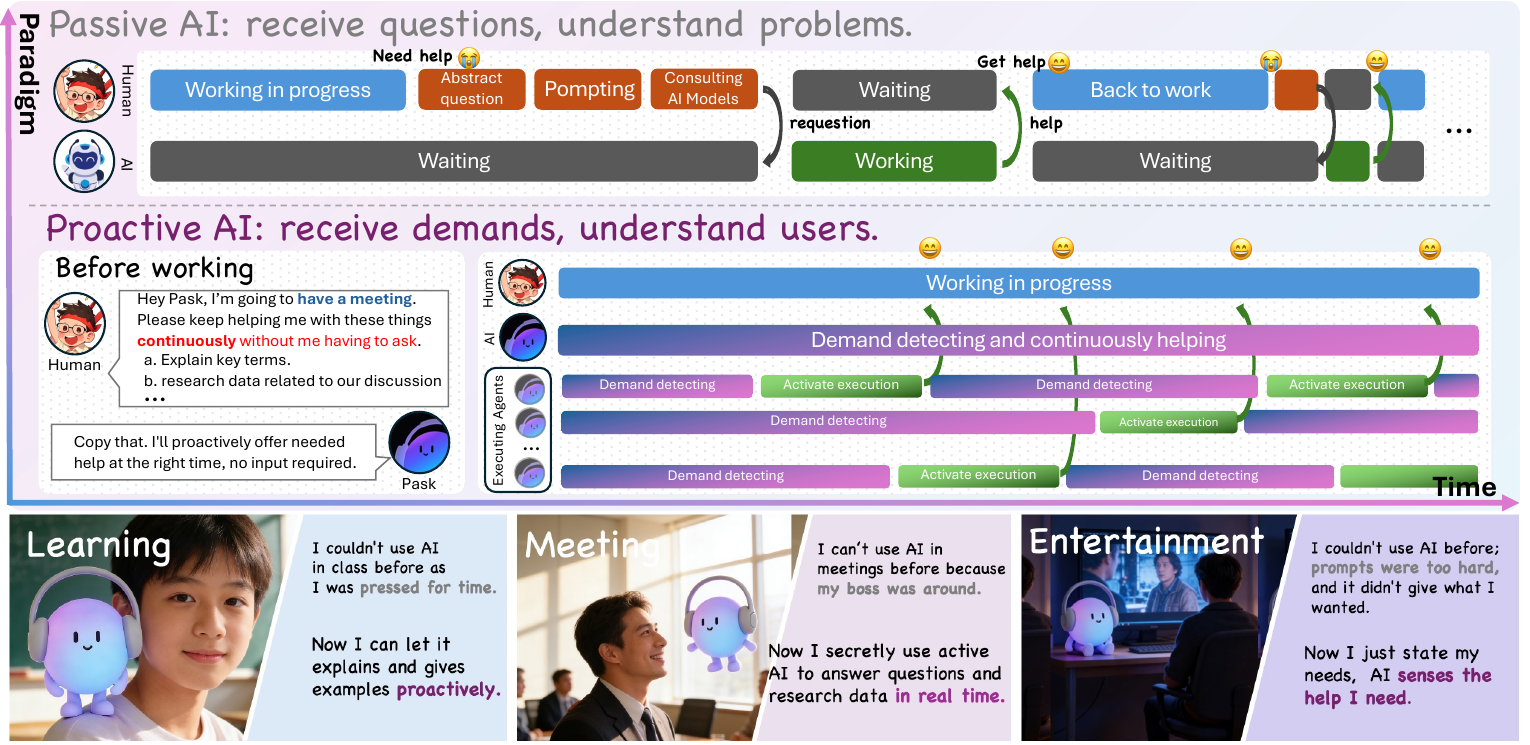}
    
    \vspace{-3mm}
    {\small \textbf{Figure 1}: Paradigm comparison of passive vs. proactive AI: continuously sensing and provides assistance proactively.}
 
\end{textblock*}
\refstepcounter{figure}

\newpage

\tableofcontents

\newpage

\input{pasksections/1-introduction}

\input{pasksections/2-DD-MM-SAS}

\input{pasksections/3-Demand-Detection}

\input{pasksections/4-Memory}

\input{pasksections/5-PAS}

\input{pasksections/6-experiments}

\input{pasksections/7-relatedWork}

\input{pasksections/8-Conclusion}

\bibliographystyle{plainnat}
\bibliography{main}

\end{document}

%% file: pasksections/1-introduction.tex
\section{Introduction}
The long-standing goal of Artificial General Intelligence (AGI) is to build systems with broad, human-level intelligence that can perceive, reason, and act in the open world~\citep{openai_gpt54, google_gemini,anthropic_claude, qwen35_2026, qwen3-omni}. Recent progress has pushed this goal forward from several directions. Reasoning-oriented models have improved deliberative ability, unified foundation models are reducing the gap across modalities, and agentic systems are bringing planning, execution, and adaptation into a single loop~\citep{mirothinker, mistral_models, qwen3-coder, cohere_commandA}. At the same time, emerging AI skills suggest more transferable and compositional capabilities beyond narrow task-specific behavior~\citep{openclaw_2025}. However, most current systems still operate in a ``you ask, I answer'' mode. We argue that this creates two basic limitations: a mismatch with how intelligence is used in the real world and an information bottleneck that prevents AI from building a deeper understanding of humans.

From the application side, real-world intelligence is constrained by timing, context, and human factors. As captured by the proverb ``Tian shi, di li, ren he,'' current AI interaction patterns often break down in practical settings. Under strict time constraints, such as watching a video or taking part in a live conversation, users often cannot stop and query an AI system (\textit{Tian shi}). In context-sensitive situations, such as meetings or social settings, invoking AI may be awkward or disruptive (\textit{Di li}). Even when AI is available, using it well still requires substantial effort: users must identify their intent, formulate a prompt, and adapt to a highly explicit and rational interaction style that many people do not naturally use (\textit{Ren he}). More broadly, if AI is to continue evolving, it must move beyond passive query-response interaction and become grounded in shared human perceptual experience. This shift would turn AI from a reactive tool into a system that can anticipate user needs and improve through a closed-loop data flywheel.

These limitations suggest that scaling model capability alone is not enough. As shown in \textbf{Figure} \textcolor{seedblue}{\textbf{1}}, proactive AI is emerging as an alternative interaction paradigm in which models perceive context in real time and offer timely assistance, shifting from reactive responders to active participants. Recent studies have explored this direction in specific domains, including programming assistance~\citep{proactive_program}, computer-operation assistance~\citep{proactive_computer}, and collaborative gameplay~\citep{proagent_gaming}. However, existing work is still focused on narrow scenarios and is mostly evaluated in controlled settings, with \textbf{\textit{limited treatment of generalization and of key real-world requirements such as interaction depth, real-time responsiveness, and robustness in dynamic environments}}. More importantly, current systems \textbf{\textit{do not yet provide an evolving memory mechanism that can accumulate long-term user understanding and adapt with the user over time}}.

\begin{tcolorbox}[
    colback=blue!5,
    colframe=blue!50!black,
    boxrule=0.6pt,
    arc=1.5mm,
    left=3mm,
    right=1.5mm,
    top=1mm,
    bottom=1mm
]
\textbf{Overall, we identify four unresolved challenges:}
\begin{enumerate}
    \item how to define a general and potentially unifying paradigm for proactive AI;
    \item how to realize its core capability, namely low-latency and accurate detection of latent user needs under continuous real-time inputs;
    \item how to equip proactive agents with evolving memory so that they can accumulate user understanding, adapt over time, and support long-term human--AI co-evolution beyond traditional chatbot systems;
    \item how to build a robust system with stable performance and low latency, so that proactive AI can work reliably in real-world applications and support continual improvement.
\end{enumerate}
\end{tcolorbox}

In this work, we introduce \textbf{Pask}, a proactive AI system designed as a complete stack rather than a set of disconnected modules. Our main argument is that proactive intelligence should be studied across four levels together: paradigm, core capability, long-term adaptation, and system implementation. Concretely, we make four connected contributions: (1) we propose \textbf{DD-MM-PAS}, a general architecture for proactive AI; (2) within this architecture, we introduce \textbf{IntentFlow}, an streaming structure foundation model for real-time demand detection; (3) we design a \textbf{hybrid co-evolving memory system} for persistent user understanding across sessions; and (4) we implement these components in a \textbf{fully functional end-to-end system} that provides practical value in real-world settings.

\begin{figure}[t]
    \centering
    \includegraphics[width=1.0\linewidth]{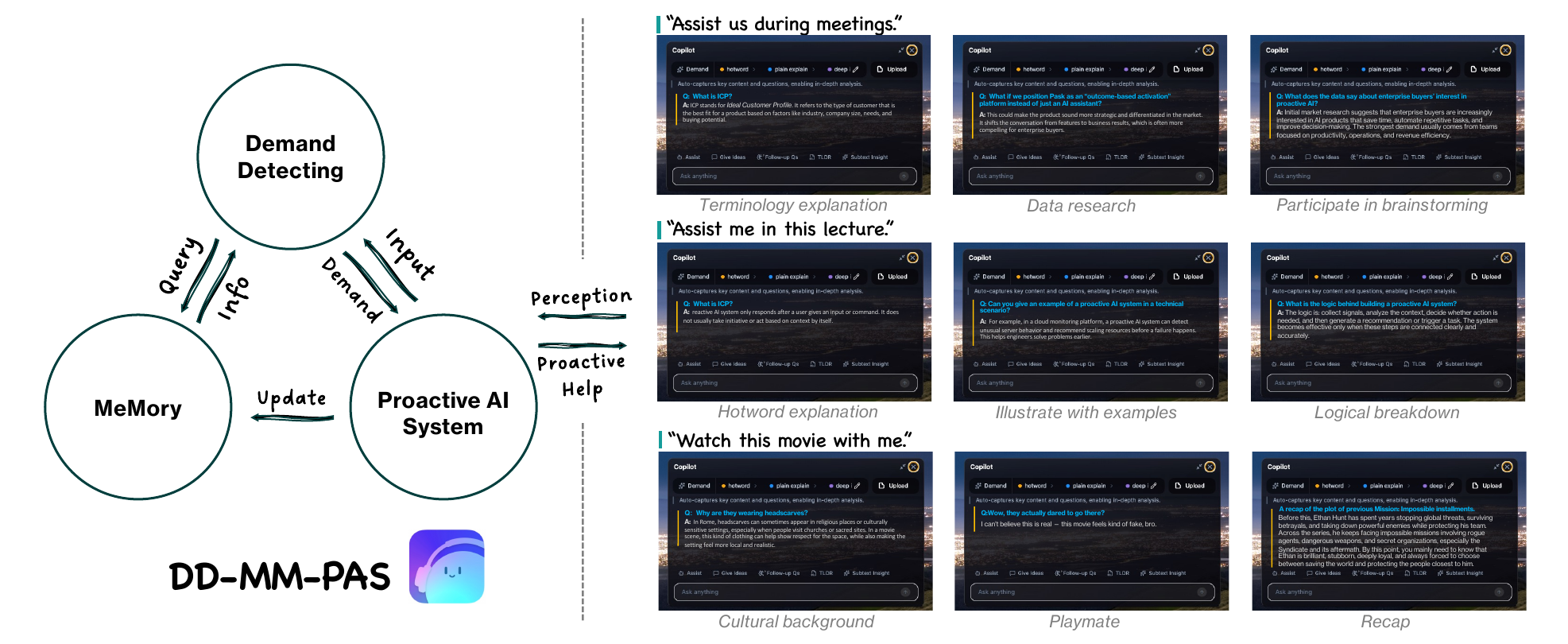}
    \caption{Illustration of the proposed DD-MM-PAS, a potentially general paradigm for proactive AI, alongside its diverse application instances. The framework is composed of three interconnected modules: Demand Detecting (DD) acts as the proactive engine to predict latent human needs from real-time signals; the Memory Module (MM) accumulates persistent, long-term data to enable deep, person-level understanding; and the Proactive Agent System (PAS) serves as the system backbone, managing real-time execution, information fusion, and memory scheduling. Together, these components transition AI from a passive, query-based responder into an active initiator of assistance across professional, academic, and daily contexts.}
    \label{fig:2-1 DDMMPAS}
\end{figure}

We first introduce DD--MM--PAS, shown in \textbf{Figure~\ref{fig:2-1 DDMMPAS}}, as a general paradigm for proactive AI with three core components. \textbf{Demand Detection (DD)} is the core proactive capability: it continuously ingests real-time signals and uses a structured user profile to infer latent user needs, allowing AI to initiate help rather than wait for requests. \textbf{Memory Module (MM)} accumulates long-term user memory over repeated use, enabling proactive AI to build person-level understanding through sustained perception and experience. This goes beyond passive, question-centric AI, which can only respond to isolated queries. \textbf{Proactive Agent System (PAS)} provides the always-on execution loop that handles information fusion, concurrent task execution, and feedback-driven updates, serving as the system backbone.

At the center of this framework is demand detection, which we view as the defining capability of proactive AI. To support it, we introduce \textbf{IntentFlow}, a fast demand detection model built specifically for proactive settings. IntentFlow takes user profiles, explicit goals, and contextual scenarios as system instructions, and processes streaming inputs continuously to decide whether and how the system should intervene as user needs evolve. To train \textbf{IntentFlow}, we build a 102k-sample dataset from both synthetic and real-world collected data through a curated pipeline. We train the model with supervised fine-tuning (SFT) followed by reinforcement learning (RL), enabling accurate demand recognition and stable decision-making under real-time conditions.

Memory is another key component of proactive AI, because long-term adaptation requires more than handling one query at a time. We therefore introduce a hybrid memory architecture that balances immediacy, completeness, and scalability. \textbf{User Memory} (similar to a cache) stores stable user traits and the most salient newly observed signals, and serves as the primary reference for demand detection. \textbf{Workspace Memory} (similar to working memory) keeps all information within a single interaction session; it is implemented through the context window of the demand detection model and is continuously organized by a dedicated memory agent. \textbf{Global Memory} (similar to external storage) is implemented as an LLM-RAG system that incrementally accumulates long-term usage data and retrieves relevant past experience when needed. Together, these three layers make memory an active mechanism for long-term human--AI co-evolution rather than a passive storage unit.

Beyond the individual components, we present proactive agent system(PAS) as a online system that integrates a user-facing frontend, a scalable server backend, and an AI backend. The system includes more than 20 models and agents and over 10 core engineering modules, providing a stable runtime environment for the DD--MM--PAS paradigm and supporting continuous deployment in real-world settings.

In summary, our contributions are fourfold:
\begin{enumerate}
    \vspace{-2mm}
    \item \textbf{A proactive AI paradigm.} We propose DD--MM--PAS, a structured paradigm for proactive AI that unifies demand detection, memory-based user modeling, and always-on agent execution in a coherent and extensible architecture.
    \vspace{-2mm}
    \item \textbf{The IntentFlow model, data pipeline, training recipe, and an open benchmark.} We introduce IntentFlow, an ultra-fast foundation model for proactive assistance, together with a data curation pipeline that produces a 102k-scale dataset from synthetic and real-world data, and a hybrid SFT--RL training recipe for accurate, low-latency demand detection under streaming inputs. We also release an open benchmark to support more rigorous and standardized evaluation in this area.
    \vspace{-2mm}
    \item \textbf{A co-evolving memory system.} We design a hybrid memory architecture that supports persistent, person-level understanding through continuous accumulation and selective retrieval of user experience, moving beyond query-centric interaction toward long-term co-evolution.
    \vspace{-2mm}
    \item \textbf{A complete, deployable system.} We present a fully functional end-to-end system, including frontend interaction, backend orchestration, and AI infrastructure, and show how proactive intelligence can be implemented stably in real-world environments.
\end{enumerate}

%% file: pasksections/2-DD-MM-SAS.tex
\section{DD-MM-PAS: A Paradigm for Proactive AI}

Proactivity is not only a desirable property of intelligent systems, but also a practical necessity in many real-world scenarios where delayed assistance already implies failure. However, existing AI systems remain largely confined to a passive regime, with limited progress on continuously inferring latent human intent and developing personalized understanding through memory over time.

We introduce DD-MM-PAS, a paradigm for Proactive AI built upon three indispensable components: \textit{DD} for demand detection, \textit{MM} for memory modeling, and \textit{PAS} for the proactive agent system. Together, these components characterize the minimal structure required for an AI system to perceive, understand, and assist prior to explicit instruction.

\subsection{Problem Formulation}

We model human--AI interaction as a continuous decision process over time. At each time step \(t\), the system observes an evolving multimodal context and must determine whether intervention is necessary, what form of assistance is appropriate, and whether such assistance justifies the risk of interruption. Unlike Passive AI, which acts only after an explicit query is provided, Proactive AI must infer latent demand directly from ongoing context and respond in a timely and calibrated manner.

The objective is to maximize the Proactive AI expected return \(J(\pi)\):
\begin{equation}
J(\pi) = \mathbb{E}_{\pi} \left[ \sum_{t=0}^{\infty} \gamma^t \left( R_{\text{help}}(\delta_t, A_t) - \lambda \cdot C_{\text{intr}}(\delta_t, A_t) \right) \right]
\end{equation}
where:
\begin{itemize}
    \item \(\delta_t\) denotes the latent user demand at time step \(t\);
    \item \(A_t\) denotes the assistance produced by the system;
    \item \(R_{\text{help}}(\delta_t, A_t)\) measures the utility of the provided assistance in satisfying the latent demand;
    \item \(C_{\text{intr}}(\delta_t, A_t)\) measures the cost of unnecessary, mistimed, or misaligned intervention;
    \item \(\lambda\) controls the trade-off between helpfulness and intrusiveness;
    \item \(\gamma\) is the temporal discount factor.
\end{itemize}

This formulation highlights that Proactive AI is fundamentally a problem of optimizing intervention under uncertainty: the system must provide useful assistance when needed, while remaining silent when no intervention is warranted. This, in turn, requires the joint support of \textit{DD}, \textit{MM}, and \textit{PAS}.

\subsection{The DD-MM-PAS Paradigm}

DD-MM-PAS decomposes proactive intelligence into three coupled functions: \textit{DD}, \textit{MM}, and \textit{PAS}. The central premise is that proactivity does not emerge from response generation alone, but from the coordinated integration of demand inference, personalized understanding, and executional capability.

\textbf{Demand Detection}\textit{(DD)} determines whether the current context implies a latent need for assistance, and directly infers the user’s underlying intent and demand. It serves as the perceptual entry point of Proactive AI, transforming continuous multimodal observations into actionable judgments about whether intervention is warranted.

\textbf{Memory Modeling}\textit{(MM)} maintains an evolving representation of the user across time. It enables the system to interpret current observations in light of accumulated personal context, thereby grounding proactive behavior in individualized understanding rather than generic pattern matching.

\textbf{Proactive Agent System}\textit{(PAS)} provides the operational substrate that turns inferred demand into effective assistance. It supports action execution through the coordinated use of external tools, computational resources, and stronger downstream models, making proactive help practically realizable.

%% file: pasksections/3-Demand-Detection.tex
\begin{figure}[!t]
    \centering
    \includegraphics[width=0.9\linewidth]{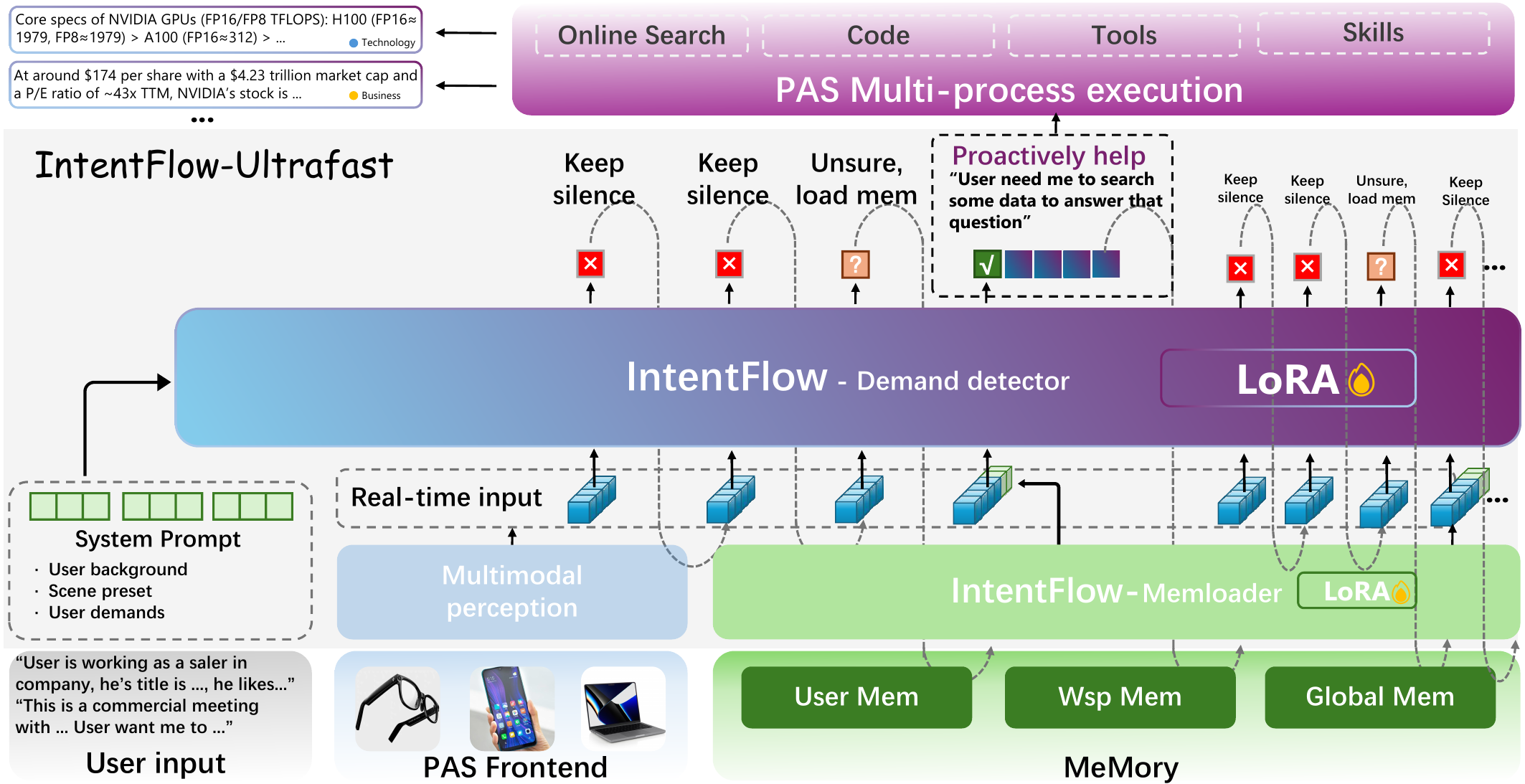}
    \caption{IntentFlow model architecture for streaming information–based human deep intent reasoning and demand detection. The system processes textual information fragments from the frontend and predicts one control token per step: \texttt{<silent>} (wait for the next message), \texttt{<fast intervention>} (low-latency understanding help, e.g., concept explanation, typically <1 s), or \texttt{<full assistance>} (invoke memory modules and global context for deep intent detection.}
    \label{fig3-1:intentflow_arch}
\end{figure}

\section{Pask-DD: IntentFlow}

In this section, we present \textbf{IntentFlow}, the core component of \textbf{Pask-DD}. Unlike conventional assistants that respond only after an explicit user query, IntentFlow is designed to anticipate a user’s assistance needs from the current information stream by aligning with their imminent intent. Our goal is not merely to generate helpful responses, but to enable large language models to infer what the user is likely to need at the current moment, conditioned on interaction context, task instructions, and memory.

We formulate IntentFlow as an end-to-end demand detection framework over textualized information streams. Given the latest information segment, the model first predicts one of three decision states: \textit{Silent}, \textit{fast intervention}, or \textit{full assistance}. These correspond to remaining inactive, providing an immediate low-latency response from the current context, and invoking memory-grounded reasoning before making a final decision. As illustrated in \textbf{Figure ~\ref{fig3-1:intentflow_arch}}, to support this process, IntentFlow adopts a dual-model architecture: the \textbf{Demand Detector} performs contextual understanding, intent prediction, and final human needs generation, while the \textbf{MemLoader} processes retrieved memory and distills relevant evidence for the detector. In the remainder of this section, we describe the architecture of IntentFlow, followed by its data curation and training procedures.

\subsection{Model Architecture}
To address accurate intent recognition and memory-grounded personalized assistance, we formulate \textbf{IntentFlow} as a real-time, turn-based primary--auxiliary architecture. The primary model, \textbf{Demand Detector}, is built on Qwen3-30B-A3B-Instruct, and the auxiliary model, \textbf{MemLoader}, is built on Qwen3-4B-Instruct. The memory extraction mechanism itself is introduced in Section~\ref{Sec4:Memory}; here we focus on the online interaction process.

\begin{figure}[!t]
    \centering
    \includegraphics[width=0.9\linewidth]{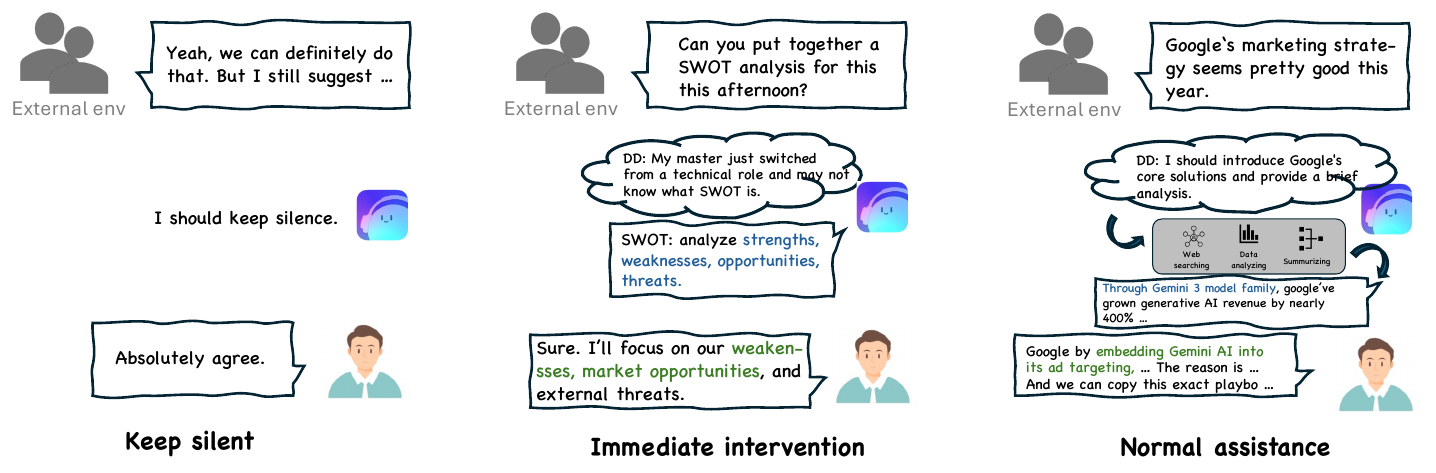}
    \caption{Different DD outcomes trigger different system modes: \textbf{<silent>} avoids unnecessary interruption, \textbf{<fast intervention>} provides immediate help, and \textbf{<full assistance>} enables agentic operations for higher-quality support.}
    \label{fig:3-2intentflow_cases}
\end{figure}

At dialogue step $t$, the system receives the latest textualized information segment $x_t$, the multi-turn interaction history $H_t=\{x_1,\dots,x_t\}$, and the external memory bank $\mathcal{M}$. The Demand Detector first predicts a decision token
\[
d_t \in \{\texttt{<silent>}, \texttt{<fast\_intervention>}, \texttt{<full\_assistance>}\},
\]
corresponding to no intervention, direct low-latency assistance from the current context, and memory-grounded reasoning, respectively. If $d_t=\texttt{<silent>}$, the system remains inactive. If $d_t=\texttt{<fast\_intervention>}$, it responds immediately from the ongoing interaction. If $d_t=\texttt{<full\_assistance>}$, the system invokes the memory pathway: the recent context is summarized into an observation summary $o_t$ and a set of salient entities $e_t$, which are used to construct a retrieval query; the retrieved evidence is then refined by MemLoader and returned to Demand Detector for final response generation or abstention. Formally,
\[
d_t = f_{\mathrm{det}}(x_t, H_t), \qquad
y_t =
\begin{cases}
\varnothing, & d_t=\texttt{<silent>},\\
f_{\mathrm{fast}}(x_t, H_t), & d_t=\texttt{<fast\_intervention>},\\
f_{\mathrm{final}}(x_t, H_t, \tilde{I}_t), & d_t=\texttt{<full\_assistance>},
\end{cases}
\]
where $\tilde{I}_t$ denotes the refined memory evidence produced by MemLoader, and $f_{\mathrm{final}}$ may either return a response or $\varnothing$ if the retrieved evidence does not justify intervention. This design casts proactive assistance as an online decision process over a growing interaction history, while flexibly coordinating direct response and memory-grounded reasoning. The resulting system behaviors under different decision modes are illustrated in \textbf{Figure~\ref{fig:3-2intentflow_cases}}.

\textbf{Silent.} When Demand Detector predicts $\texttt{<silent>}$, IntentFlow produces no output and waits for the next dialogue step. This mode avoids unnecessary interruption and preserves a natural interaction rhythm when the user state does not suggest a meaningful opportunity for assistance.

\textbf{Fast intervention.} When the model predicts $\texttt{<fast\_intervention>}$, it directly generates assistance from the latest stream and recent interaction context, without consulting external memory. This path is suitable for explicit and short-horizon needs, such as clarifying a concept, explaining an instruction, or addressing a locally resolvable request. Because the required evidence is already available in the ongoing interaction, this mode minimizes latency while maintaining responsiveness.

\textbf{Full assistance.} When Demand Detector emits $\texttt{<full\_assistance>}$, the system enters a memory-grounded reasoning pipeline for personalized proactive assistance. An LLM-based agent first processes the recent context to extract an observation summary $o_t$ and salient entities $e_t$. These signals are combined with the latest input and interaction history to form a retrieval query, which is sent to the Pask-MM module:
\[
q_t=\psi(x_t,H_t,o_t,e_t), \qquad I_t=\mathrm{PaskMM}(q_t,\mathcal{M}).
\]
Since the retrieved evidence may still contain redundancy or weakly relevant content, MemLoader further distills it into a compact set of core information,
\[
\tilde{I}_t=\mathrm{MemLoader}(I_t,x_t,H_t),
\]
which is then returned to Demand Detector for final decision making. In this mode, the model may generate a personalized response when the memory-grounded evidence supports intervention, or remain silent if deeper reasoning suggests that no assistance is necessary.

\subsection{Data Curation} 
\label{sec:data}

High-quality data on human intent is essential for training IntentFlow to directly recognize latent user needs without relying on explicit persona-style reasoning. Existing models often require an additional reasoning step to infer what a user may need, which conflicts with the low first-token latency required by proactive assistance; moreover, even with such reasoning, the inferred needs are often shallow or unreliable. To address this limitation, we follow a two-stage data-centric strategy: we first internalize this reasoning process through large-scale, high-quality supervision, so that intent understanding becomes an efficient forward prediction rather than a costly explicit deliberation step, and then further align the model with deeper human intent through reinforcement learning, enabling more accurate and nuanced proactive assistance beyond what supervised training alone can provide. 

To support this strategy, we construct \textbf{LatentNeeds}, a dataset consisting of \textbf{100k} synthetic samples for supervised fine-tuning and \textbf{2.1K} real-world sessions collected from users. Among the real-world data, \textbf{2K} sessions are used for reinforcement learning alignment, while the remaining \textbf{100} sessions are reserved for benchmark construction. An overview of the dataset composition, generation pipeline, and training strategy is illustrated in \textbf{Figure~\ref{fig:3-3 data_pipeline}}.

\subsubsection{LatentNeeds-100k for Finetuning: }  
We first construct the large-scale supervised part of our training pipeline, \textbf{LatentNeeds-100k}, to build the foundation for IntentFlow’s intent prediction capability. We define three broad scenario domains, \textbf{\textit{learning}, \textit{meetings}, and \textit{daily life}}, and further divide each domain into fine-grained subcategories that serve as the basic taxonomy for data construction. Based on this taxonomy, the dataset is built through the following pipeline: \textit{\underline{(1) Public-data grounding.}} We use tools to collect metadata from publicly available sources associated with each subcategory and extract realistic Internet content as the source material for subsequent synthesis. \textit{\underline{(2) Multi-agent information-stream generation.}} We define the relevant roles in each scenario and instantiate them as multiple agents that interact around a shared topic or event trajectory, producing realistic information streams that mimic natural communication. \textit{\underline{(3) Iterative human-intent generation.}} For each scenario, AI generates role-specific background information and, after each interaction turn, infers the receiver role’s ongoing thoughts, latent intentions, and potential needs. \textit{\underline{(4) Reformatting.}} The resulting data are reorganized for each role into structured tuples of the form $\{b_i,\, c_i,\, (x_i, d_i, y_i)\}$, where $b_i$ denotes the human background, $c_i$ the context, $x_i$ the observed information, $d_i$ the corresponding inferred decision state, and $y_i$ the target assistance content for the $i$-th sample. \textit{\underline{(5) AI-based post-filtering.}} A final LLM-based review stage evaluates the quality of each sample and determines whether it should be retained. The resulting dataset encodes complete first-person intent inference together with explicit role background and memory, providing scalable supervision for training models to predict human needs directly from ongoing information streams.

\subsubsection{LatentNeeds-2K for Intent-alignment Reinforcement Learning: }  
Reinforcement learning is known for delivering substantial gains from limited data, highlighting its strong data efficiency. Motivated by this property, we construct \textbf{LatentNeeds-2K} from highly curated real-world data to align IntentFlow with human intent under realistic conditions. The data are drawn from two sources: one consists of real user sessions, collected with user consent and anonymized before use; the other is sourced from the Internet to supplement scenarios that are underrepresented in the real-user data. After collection, we invite the information owners or relevant domain participants to refine the annotated demands through deletion, addition, and editing, ensuring that the final targets better reflect genuine human needs.

In total, we collaborate with \textbf{143} users to collect \textbf{2.1K} real-world interaction sessions. Among them, \textbf{2K sessions} are segmented and used as the reinforcement learning dataset for aligning IntentFlow with realistic human intent, while the remaining \textbf{100 sessions} are reserved for constructing our evaluation benchmark, introduced in Section~\ref{sec:latentneeds}.

\begin{figure}[!t]
    \centering
    \includegraphics[width=1.0\linewidth]{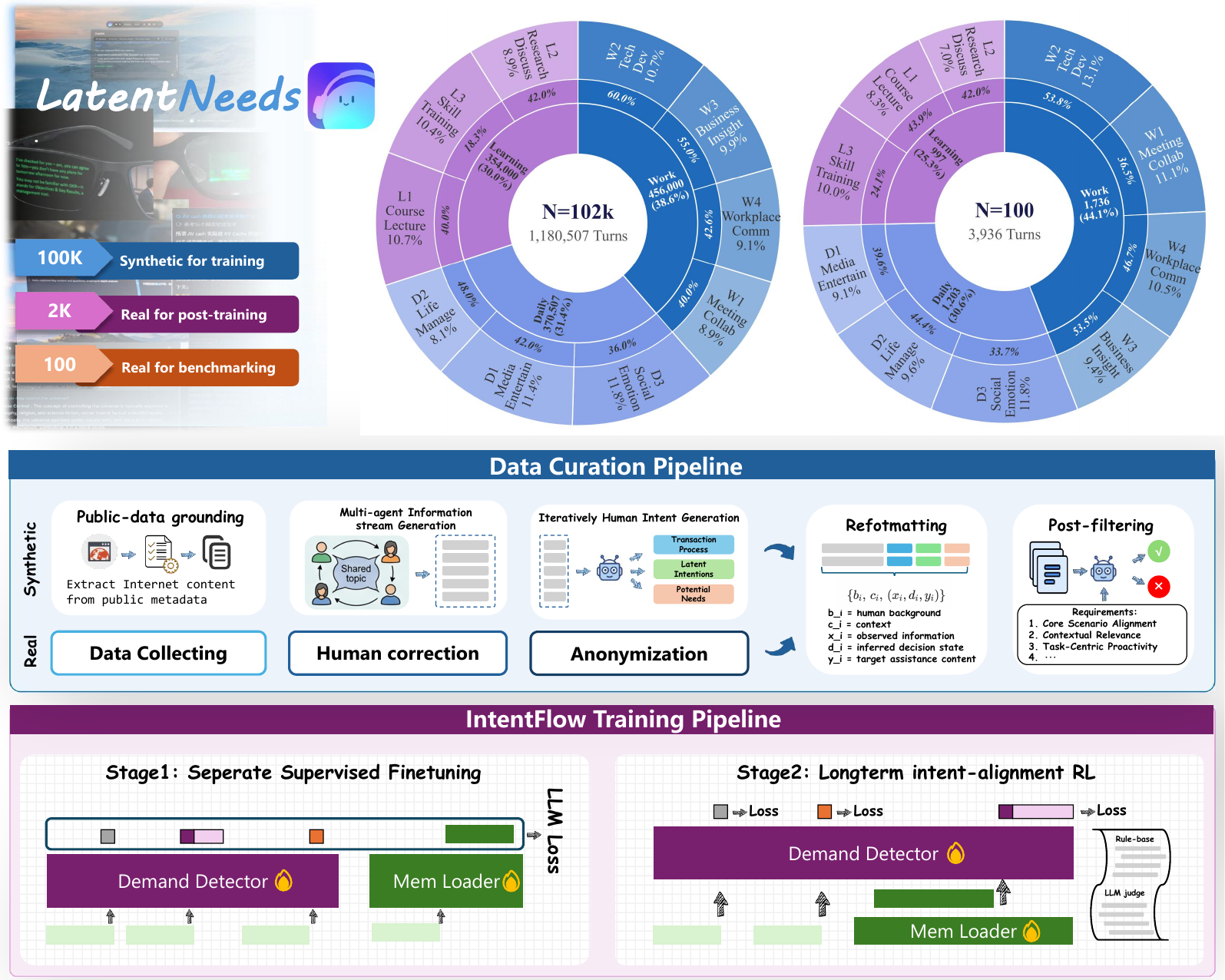}
    \caption{Overview of the LatentNeeds dataset and benchmark composition, showing the distribution across three major domains (meetings, learning, and daily life) and their corresponding subcategories. The middle part summarizes the data construction pipeline, including public-data grounding, multi-agent information-stream generation, iterative human-intent generation, structured reformatting, and AI-based post-filtering, together with the collection and refinement of real-world sessions. The bottom part outlines the training strategy: separate supervised fine-tuning of the Demand Detector and MemLoader, followed by joint reinforcement learning for deeper human-intent alignment.}
    \label{fig:3-3 data_pipeline}
\end{figure}

\subsection{Training Procedure}
We adopt a two-stage training strategy (\textbf{Figure~\ref{fig:3-3 data_pipeline}}), consisting of separate supervised fine-tuning and reinforcement learning for deeper intent alignment.

\subsubsection{Separate supervised fine-tuning.}

We first train \textbf{Demand Detector} and \textbf{MemLoader} independently using \textbf{LatentNeeds-100k}, with supervision targets tailored to their respective roles. For \textbf{Demand Detector}, the input consists of the human background $b$, the interaction context $c$, the information history $x_{1:t}$, and the previous decision sequence $d_{1:t-1}$, and the model is trained to predict both the current decision token $d_t$ and the corresponding assistance content $y_t$. Its training objective is defined as $\mathcal{L}_{\mathrm{det}} = \mathcal{L}(d_t, y_t \mid b, c, x_{1:t}, d_{1:t-1})$. To improve robustness over long interaction horizons, we adopt a curriculum over dialogue length and progressively expand the training horizon in three stages until reaching 15 turns. For \textbf{MemLoader}, the input consists of the human background $b$, the interaction context $c$, the information history $x_{1:t}$, and the retrieved memory candidates $I_t$ returned by the Pask-MM module, and the model is trained to produce the curated memory output $\tilde{I}_t$ for the current step, with objective $\mathcal{L}_{\mathrm{mem}} = \mathcal{L}(\tilde{I}_t \mid b, c, x_{1:t}, I_t)$. This independent supervised fine-tuning stage establishes the basic capabilities of intent detection, demand generation, and memory distillation before reinforcement learning for deeper intent alignment.

\subsubsection{Reinforcement learning for deep human intent alignment.}

We further apply reinforcement learning on \textbf{LatentNeeds-2K}, which is constructed from real-world data, to extend IntentFlow toward deeper alignment with realistic human intent. At this stage, we jointly optimize \textbf{Demand Detector} and \textbf{MemLoader}, allowing the full system to adapt both its intervention decisions and memory-grounded assistance under realistic interaction dynamics. Our reinforcement learning stage adopts the \textbf{DAPO} setting.

\noindent$\cdot$ \textbf{Rule-based reward:} We use rule-based rewards for aspects that can be verified precisely. Specifically, the reward checks whether the model outputs a valid special token and a well-formed demand, and whether the intervention happens at an appropriate time. These signals provide stable supervision for both output format and intervention timing.

\noindent$\cdot$ \textbf{Model-based reward:} For aspects that do not admit explicit rules, we adopt an LLM-as-a-judge protocol. The evaluator scores the generated demand on three dimensions: \emph{alignment}, measuring consistency with the reference need; \emph{reasonableness}, measuring whether the demand is plausible under the current context; and \emph{necessity}, measuring whether the intervention is genuinely useful rather than redundant. Each dimension is scored from 1 to 5, and the model-based reward is computed as
\[
R_{\mathrm{model}} = w_1 s_{\mathrm{align}} + w_2 s_{\mathrm{reasonable}} + w_3 s_{\mathrm{necessary}},
\]
where $s_{\mathrm{align}}$, $s_{\mathrm{reasonable}}$, and $s_{\mathrm{necessary}} \in \{1,\dots,5\}$ are the three scores, and $w_1$, $w_2$, and $w_3$ are their corresponding weights. The final reward is the sum of the rule-based and model-based rewards.

%% file: pasksections/4-Memory.tex
\section{Pask-MM: Self-Evolving Hierarchical Memory Modeling}\label{Sec4:Memory}

A Proactive AI system must precisely anticipate human needs to provide non-intrusive assistance. To achieve high-fidelity cognitive profiling, its foundation lies in a dynamic, self-evolving memory architecture. In this section, we introduce PASK-MM, the core memory module of the Pask system. Our objective is to accommodate four fundamental challenges in proactive memory modeling: i) processing massive continuous token streams from long-term real-time inputs; ii) operating under a strict latency constraint of at most one second; iii) ensuring high decision accuracy to avoid disruptive interventions; and iv) enabling continual, scalable evolution based on long-horizon interaction data without triggering compute explosion.

To satisfy these constraints, we draw inspiration from classical computer architecture and design PASK-MM as a hierarchical ``Cache--Main Memory--External Storage'' system. A key design feature is a bounded tree-structured representation spanning from coarse contextual abstractions to fine-grained semantic records. To physically reconcile the contradiction between multi-level tree traversal and sub-second latency, we decouple immediate state reasoning from deep historical retrieval. The system utilizes an asynchronous coarse-to-fine traversal combined with Retrieval-Augmented Generation (RAG). Finally, we introduce a bounded self-evolution strategy featuring conflict resolution, memory decay, and lazy merging, ensuring that the memory system maintains a steady and compact state throughout long-term deployments.

\begin{figure}[t]
    \centering
    \includegraphics[width=1\linewidth]{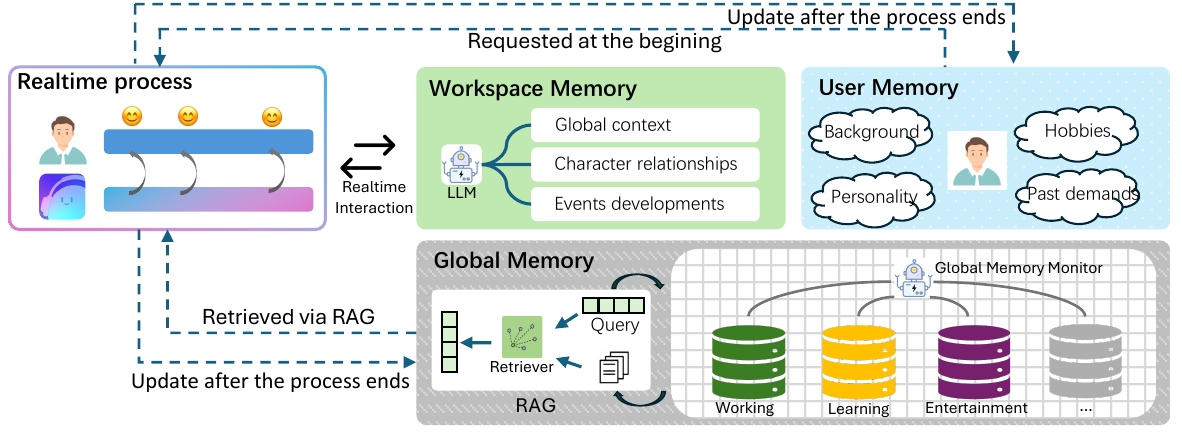}
    \caption{Overview of the internal architecture of \textbf{Pask-MM}. Inspired by hierarchical memory designs in modern agent systems, the framework organizes memory into three functional components with different access frequencies: user memory $\mathcal{M}_{\text{user}}$, workspace memory $\mathcal{M}_{\text{wsp}}$, and global memory $\mathcal{M}_{\text{global}}$. $\mathcal{M}_{\text{user}}$ stores stable user profiles for personalization, $\mathcal{M}_{\text{wsp}}$ maintains short-term session dynamics for real-time inference, and $\mathcal{M}_{\text{global}}$ records long-term episodic knowledge for retrieval and validation. During interaction, workspace memory is updated continuously to support low-latency reasoning, while $\mathcal{M}_{\text{user}}$ and $\mathcal{M}_{\text{global}}$ evolve only after session completion through offline maintenance and compression.}
    \label{fig:4-1 Pask-MM arch}
\end{figure}

\subsection{Architecture Definition}
The architecture of Pask-MM is driven by the dual imperatives of high precision and real-time responsiveness. We categorize the framework into three distinct components based on their functional access frequency: user memory ($\mathcal{M}_{\text{user}}$), workspace memory ($\mathcal{M}_{\text{wsp}}$), and global memory ($\mathcal{M}_{\text{global}}$). The framework is illustrated in \textbf{Figure~\ref{fig:4-1 Pask-MM arch}}

\paragraph{$\mathcal{M}_{\text{user}}$ (User Memory as Cache)}
User background dictates the paradigm of proactive assistance. We design $\mathcal{M}_{\text{user}}$ as a high-priority, dense cache directly injected into the system prompt. It is a strictly controlled profile representation, summarizing stable attributes (e.g., domain expertise, personalized thresholds, and behavioral priors). By leveraging KV-cache reuse, $\mathcal{M}_{\text{user}}$ provides an immediately accessible zero-latency user model for inference.

\paragraph{$\mathcal{M}_{\text{wsp}}$ (Workspace Memory as Main Memory)}
$\mathcal{M}_{\text{wsp}}$ maintains session-level local dynamics during an active interaction lifecycle. Analogous to main memory, it is continuously updated alongside active reasoning processes. $\mathcal{M}_{\text{wsp}}$ buffers the conversational history, intermediate environmental variables, and internal system states, ensuring the agent anchors its immediate processing to the ongoing task while preserving short-horizon temporal coherence.

\paragraph{$\mathcal{M}_{\text{global}}$ (Global Memory as External Storage)}
$\mathcal{M}_{\text{global}}$ stores the comprehensive interaction history, utilized for validating hypotheses, retrieving episodic knowledge, and tracking long-term projects. $\mathcal{M}_{\text{global}}$ is formalized as a rooted tree $\mathcal{T}=(\mathcal{V},\mathcal{E})$. Each internal node $v\in\mathcal{V}$ stores a semantic tag $\mathrm{tag}(v)$ representing a hierarchical abstraction. Each leaf node $v_{\text{leaf}}\in\mathcal{V}$ carries a payload
\begin{equation}
    \mathrm{payload}(v_{\text{leaf}})=m=\langle t_m, c_m \rangle,
\end{equation}
where $t_m$ denotes the fine-grained tag and $c_m$ denotes the episodic content. This topology simultaneously supports scalable macro-summarization and precise micro-retrieval via RAG.

\subsection{Memory Access and Inference Mechanism}
The core challenge of the access mechanism is to extract relevant contexts while guaranteeing an ultra-low latency of $\le 1$ second. To achieve this, Pask employs a decoupled state-return and asynchronous retrieval strategy.

At each time step $t$, the input to the memory module is formalized as

\vspace{-4mm}
\begin{equation}
    q_t = \langle o_t, H_t, e_t, \tau_t \rangle,
\end{equation}

\vspace{-2mm}
where $o_t$ denotes the raw observation, $H_t=\{x_1,\dots,x_t\}$ denotes the session history up to time $t$, $e_t$ denotes the extracted entities, and $\tau_t$ denotes the temporal metadata.

To ensure rapid response, the system immediately relies on the active workspace memory from the previous step:
\begin{equation}
    \mathcal{M}_{\text{wsp}}^{t-1} = \langle g_{t-1}, l_{t-1}, s_{t-1}, i_{t-1} \rangle,
\end{equation}
where $g_{t-1}$ and $l_{t-1}$ are the global and local session contexts, $s_{t-1}$ is the internal memory state, and $i_{t-1}$ records the interaction trace maintained in workspace memory. The workspace state is updated by

\vspace{-4mm}
\begin{equation}
    s_t = \mathrm{TrackState}(s_{t-1}, o_t, H_t, \tau_t).
\end{equation}

\vspace{-2mm}
Accordingly, the lightweight workspace return is defined as

\vspace{-4mm}
\begin{equation}
    I_t^{\text{wsp}} = \langle g_{t-1}, l_{t-1}, s_t, i_{t-1} \rangle.
\end{equation}

\vspace{-2mm}
Simultaneously, to resolve the latency conflict of deep retrieval, access to $\mathcal{M}_{\text{global}}$ is executed as an \textit{asynchronous lazy evaluation}. Instead of performing a full tree traversal at every step, the system reuses the previously located node $v_k^{t-1}\in\mathcal{V}$ as the anchor of the current event. Its semantic tag serves as the coarse-grained event memory:
\begin{equation}
    I_t^{\text{global}} = \mathrm{tag}\!\left(v_k^{t-1}\right).
\end{equation}
Conditioned on this anchor, fine-grained evidence is retrieved only from the descendants of $v_k^{t-1}$ through localized RAG:

\vspace{-5mm}
\begin{equation}
    I_t^{\text{rag}} = \mathrm{RAG}\!\left(\left\{ c_m \,\middle|\, v_{\text{leaf}} \in \mathrm{Desc}\!\left(v_k^{t-1}\right),\ \mathrm{payload}(v_{\text{leaf}})=\langle t_m,c_m\rangle \right\}, q_t\right).
\end{equation}

\vspace{-2mm}
In this way, $I_t^{\text{global}}$ provides a stable coarse memory of the ongoing event, while $I_t^{\text{rag}}$ supplies fine-grained episodic evidence from the corresponding subtree without introducing full-tree retrieval overhead.

All validated signals are then unified into the final inference representation:
\begin{equation}
    I_t = \langle \mathcal{M}_{\text{user}}, I_t^{\text{wsp}}, I_t^{\text{global}}, I_t^{\text{rag}} \rangle.
\end{equation}

\subsection{Memory Self-Evolution and Maintenance}
Continuous interaction inevitably introduces information conflicts, habit shifts, and data explosion. Therefore, ``evolution'' must transcend linear appending. Triggered strictly post-session (offline) to ensure zero impact on inference latency, Pask implements a structured maintenance protocol addressing conflict resolution, memory decay, and structural compression.

\paragraph{Evolution of $\mathcal{M}_{\text{user}}$: Conflict Resolution and Forgetting}
Given a newly terminated request trajectory $Q = \{ q_1, \dots, q_n \}$, the system extracts candidate user traits $\mathcal{U}'$. To prevent sudden input errors from corrupting stable profiles, and to account for natural habit drift, $\mathcal{M}_{\text{user}}$ applies a time-decayed Bayesian update:
\begin{equation}
    \mathcal{M}_{\text{user}}^{(T)} = \mathrm{Decay}(\mathcal{M}_{\text{user}}^{(T-1)}, \Delta \tau) \oplus \mathrm{ResolveConflict}(\mathcal{U}', \mathcal{M}_{\text{user}}^{(T-1)}).
\end{equation}
Here, $\mathrm{Decay}(\cdot)$ gradually lowers the confidence weights of outdated preferences over time interval $\Delta \tau$. $\mathrm{ResolveConflict}(\cdot)$ compares new evidence $\mathcal{U}'$ against existing traits: reinforcing matched items, explicitly overwriting decayed contradictions, and discarding low-confidence anomalies.

\paragraph{Evolution of $\mathcal{M}_{\text{global}}$: Lazy Merging and Bounded Scaling}
A naive bottom-up tree update (updating all ancestors upon every leaf insertion) incurs catastrophic $O(N \log N)$ compute overhead over long horizons. To resolve this scalability crisis, Pask employs a \textit{Lazy Merging and Bounded-Depth} strategy.

New episodes $Q$ are initially inserted as raw leaves under a local buffer node $v_{\text{buffer}}$:
\begin{equation}
    \mathcal{V} \leftarrow \mathcal{V} \cup \{ v_{\text{new}} \}, 
    \qquad
    \mathcal{E} \leftarrow \mathcal{E} \cup \{ (v_{\text{buffer}}, v_{\text{new}}) \}.
\end{equation}
Here, $v_{\text{new}}$ is a newly created leaf node whose payload stores the memory item extracted from $Q$. Ancestor nodes are \textit{not} immediately recomputed. Instead, a background compression is triggered only when the child count of a parent node reaches a threshold $\eta_{\mathrm{merge}}$:
\begin{equation}
    \text{if } |\mathrm{Children}(v_p)| > \eta_{\mathrm{merge}},
    \qquad
    v_p^* = \mathrm{CompressAndMerge}(\mathrm{Children}(v_p)).
\end{equation}
During this operation, redundant sibling nodes are deduplicated and obsolete information is pruned. Furthermore, the tree is constrained to a maximum depth $D_{\max}$.

Through this dual mechanism, although localized fine-grained leaves accumulate continuously at the bottom, the upper-level hierarchical topology remains highly compact. This fundamentally resolves the storage maintenance crisis, allowing the system to maintain a steady architectural state and sustain bounded retrieval latency regardless of the lifecycle duration.

%% file: pasksections/5-PAS.tex
\begin{figure}[t]
    \centering
    \includegraphics[width=0.9\linewidth]{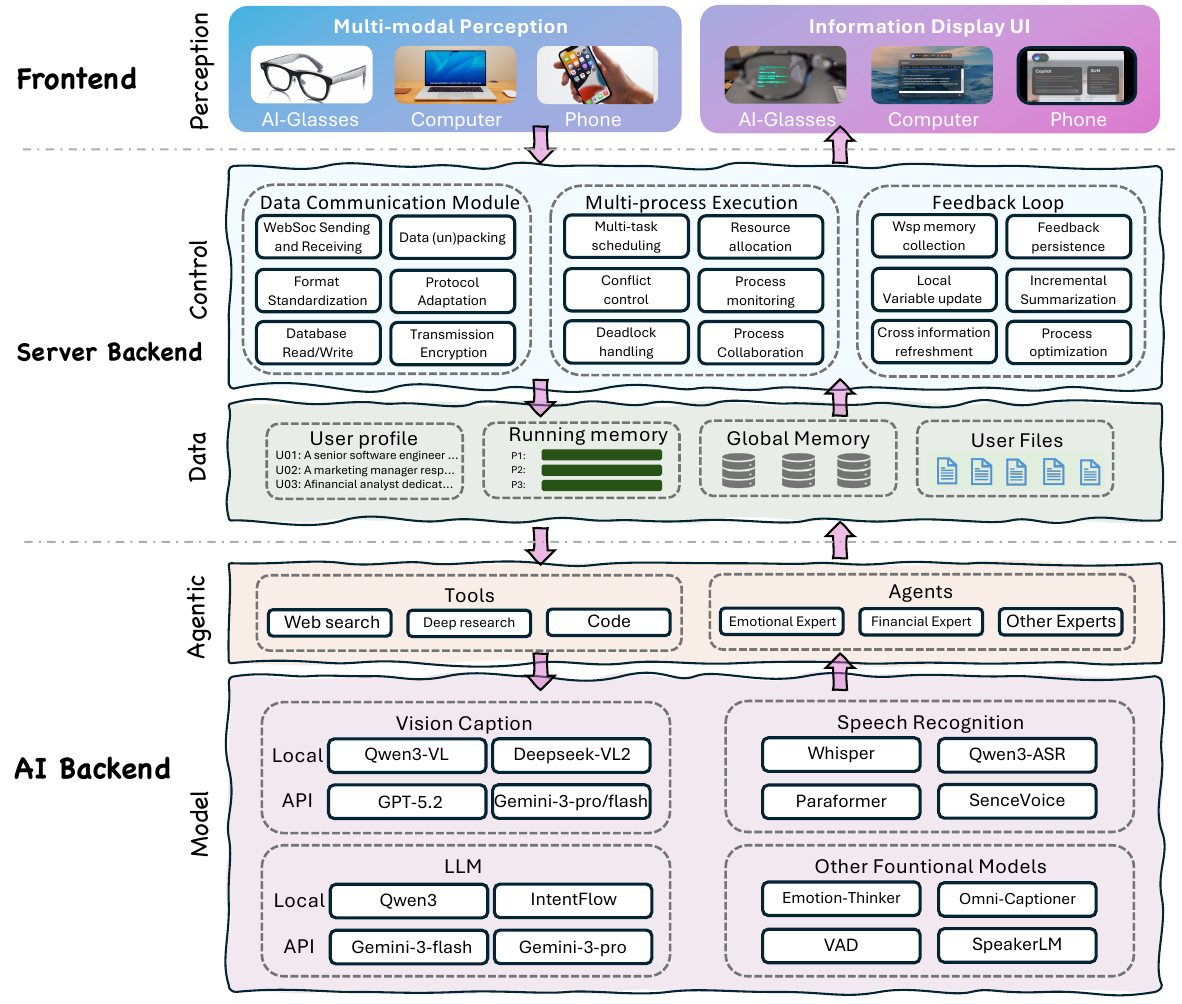}
    \caption{System architecture of \textbf{Pask-PAS}, illustrating how frontend devices, server infrastructure, and AI backends form an always-on loop for perception, understanding, and proactive action.}
    \label{fig:5-1 PASK-PAS architecture}
\end{figure}

\section{Pask-PAS: System Implementation}

In this section, we introduce the components of \textbf{Pask-PAS}. From hardware interfaces and runtime infrastructure to memory, agentic execution, and frontier models, \textsc{PAS} connects the full system stack into an always-on loop for perception, understanding, and action. The overall structure is illustrated in Figure~\ref{fig:5-1 PASK-PAS architecture}.

\textbf{Frontend Layer.} connects PAS to the devices that most naturally carry human context---AI glasses, computers, and phones---so the system can directly share the user’s perceptual stream.

\textbf{Server Backend.} provides the runtime foundation for stable coordination and memory management. \textbf{Control Layer.} keeps PAS stable as an always-on multi-process runtime through communication, scheduling, conflict isolation, resource coordination, and feedback circulation. \textbf{Data Layer.} provides the foundation of Pask-MM, implementing user profile, running memory, global memory, and user files through production-grade data infrastructure such as Redis-style hot-state management and object-store / vector-store backed long-term persistence.

\textbf{AI Backend.} gives PAS agentic intelligence, enabling it not only to perceive and respond, but to proactively do useful work through tool use and expert policies. Here, demand detection is only the bridge---IntentFlow decides what human intentially need, while the actual help is carried out by a frontier model pool, including but not limited to \textbf{Vision Captioning} models such as Qwen3-VL~\citep{qwen3-vl}, DeepSeek-VL2~\citep{deepseek-vl2}, GPT-5~\citep{openai_gpt54}, and Gemini3-pro~\citep{google_gemini};\textbf{ Speech Recognition} models such as Whisper~\citep{whisper}, Qwen3-ASR~\citep{qwen3-asr}, Paraformer~\citep{paraformer}, and SenseVoice~\citep{sensevoice}; \textbf{LLM} models such as Qwen3~\citep{qwen35_2026}, IntentFlow, and Gemini3-flash~\citep{google_gemini}; and Other Foundational Models such as Emotion-Thinker\cite{emotionthinker}, Omni-Captioner\cite{omni-captioner}, VAD(voice activity detection), and SpeakerLM~\citep{speakerlm}.In this work, we explore proactive AI in more realistic settings, where useful assistance depends on inferring latent user needs from ongoing context and leveraging evolving user memory. To support this setting, we present Pask, a proactive AI system that integrates the DD–MM–PAS paradigm, the IntentFlow model for demand detection, and a three-level memory module for longer-term personalization. We also introduce LatentNeeds-Bench as a benchmark for studying proactive assistance under real-world conditions. Experimental results suggest that, under latency constraints, IntentFlow can achieve competitive performance while in some cases identifying user intents that are less explicit or more deeply contextualized.

Overall, our findings indicate that proactive assistance may be a promising direction for moving beyond purely reactive AI interaction. Rather than viewing proactivity as a standalone capability, this work highlights the potential value of studying demand detection, memory, and system design together in a unified framework. We hope Pask and the accompanying benchmark can provide a useful basis for future research on proactive AI and long-term human–AI interaction.

%% file: pasksections/6-experiments.tex


\vspace{5mm}
\begin{figure*}[t]
\centering
\includegraphics[width=\linewidth]{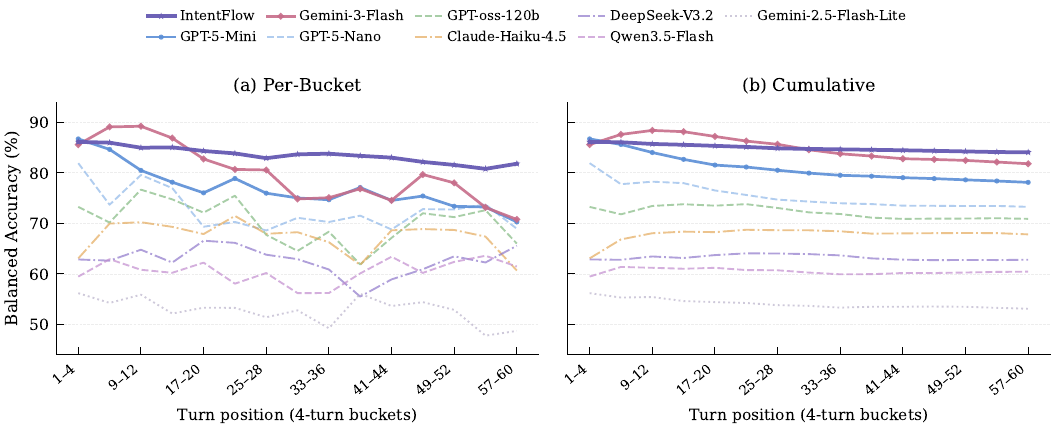}
\caption{Balanced accuracy as a function of conversation depth.
Turns are grouped into non-overlapping buckets of 4 consecutive turns (1--4, 5--8, \ldots, 57--60).
\textbf{(a)}~Per-bucket accuracy: each point is the balanced average of demand recall and non-demand precision within that bucket alone, revealing local fluctuations.
\textbf{(b)}~Cumulative accuracy: each point aggregates all turns from the start up to and including the current bucket, yielding a smoothed view of overall degradation.
IntentFlow maintains $>$80\% balanced accuracy across all buckets, while the strongest baseline (Gemini-3-Flash) drops from 85.6 to 70.8 ($\Delta$=--17.3\%).}
\label{fig:multi_turn_combined}
\end{figure*}

\section{Experiments}

\input{pasktables/main_result}

\subsection{Experimental Setup}
\label{sec:exp_setup}

\subsubsection{LatentNeeds-Bench}
\label{sec:latentneeds}

\textbf{Benchmark.} We evaluate on \textsc{LatentNeeds-Bench}, a multi-turn proactive demand detection benchmark built from real speech transcriptions (see \S\ref{sec:data}). The evaluation split contains 100 sessions (3{,}936 turns), evenly distributed across 10 subcategories from three domains: Work (W1--W4), Learning (L1--L3), and Daily (D1--D3), with 10 sessions per subcategory. Each turn is annotated with a binary demand label; demand turns additionally include a demand type (\textit{Requirement} or \textit{Insight}) and a reference response. The overall demand rate is 41.9\%.

\textbf{Protocol.} At each turn, the model is given the same input: a system prompt defining its role as a proactive assistant for a designated primary user, scene context (characters and setting), an optional memory summary of earlier conversation, and the full dialogue history up to the current turn. The model must either generate a concise proactive response or output \texttt{[NO\_DEMAND]}. To reduce prompt sensitivity, we evaluate each model under three prompt levels---\textit{encouraging}, \textit{neutral}, and \textit{suppressing}---which progressively raise the intervention threshold. We report the best-performing level for each model. All models use temperature 0.3.

\textbf{Scoring.} We use a hybrid scheme combining rule-based cases and a two-round LLM-as-judge protocol with GPT-5.2. If both the annotation and the model indicate no demand, the turn is counted as a true negative; if the model stays silent on a demand turn, it is counted as a false negative. For all turns where the model produces a response, the judge first evaluates the current turn, the ground-truth annotation, and the model response, and outputs \texttt{1}, \texttt{0}, or \texttt{NEED\_CONTEXT}; if needed, a second round provides the full dialogue history for a final binary decision. Our primary metric is balanced accuracy:
\[
\text{Balanced Accuracy}
= \frac{1}{2}
\left(
\frac{TP}{TP+FN}
+
\frac{TN}{TN+FP}
\right),
\]
which equally weights accuracy on demand and non-demand turns.



\subsubsection{Baselines}

IntentFlow is built on top of the \textbf{Qwen3-30B-A3B} base model. In terms of overall capability, this model is roughly on par with the current state-of-the-art open-source LLMs. Therefore, for a fair comparison within the open-source ecosystem, we mainly select two representative strong baselines: \textbf{GPT-oss-120B}~\citep{openai_gpt_oss_120b} and \textbf{DeepSeek-V3.2}~\citep{deepseekv3.2}. Both models are widely recognized as competitive open-weight models with strong reasoning and general instruction-following ability, and thus provide a meaningful reference point for evaluating the effectiveness of our proactive modeling approach.

Besides open-source models, we also include several widely used \textbf{closed-source commercial models} as additional baselines. These models represent commonly deployed systems in real-world applications and provide another perspective on the relative performance of our approach. Specifically, we evaluate against \textbf{GPT-5-Mini}~\citep{openai_gpt5_mini}, \textbf{GPT-5-Nano}~\citep{openai_gpt5_nano}, \textbf{Gemini-3-Flash}~\citep{google_gemini_3_flash}, \textbf{Gemini-2.5-Flash-Lite}~\citep{google_gemini_25_flash_lite}, \textbf{Claude-Haiku-4.5}~\citep{anthropic_claude_haiku_45}, and \textbf{Qwen3.5-Flash}~\citep{qwen35_2026}. These models are generally designed for fast response and cost-efficient deployment, and are commonly used in production environments where latency and throughput are important considerations.

\subsection{Main Result}

Table~\ref{tab:main_results} studies the simplest form of intent-demand detection: given a single user fragment and a user profile, the model must decide whether help is needed and, if so, what kind. This setting is \textbf{less about deep reasoning} than about \textbf{recovering a user’s simple latent need from minimal context}. From the results, we draw three main findings.

\textbf{\textcolor{sbase03}{[ Language models remain weak at this task. ]}} Even with prompts that encourage proactive assistance, many models perform poorly on the Demand split, including Gemini-2.5-Flash-Lite (18.8), Qwen3.5-Flash (29.1), DeepSeek-V3.2 (32.3), and Claude-Haiku-4.5 (38.9). Stronger models improve substantially, but the task is still \textbf{far from solved}: GPT-5-Mini reaches 66.5, GPT-5-Nano 71.2, and only Gemini-3-Flash exceeds 83 points (83.3). This suggests that the main bottleneck is \textbf{not complex reasoning}, but \textbf{reliably identifying a user’s unstated yet simple need}. After training, our model reaches 83.1 on Demand, \textbf{essentially matching Gemini-3-Flash} and clearly outperforming GPT-5-Mini, GPT-5-Nano, and GPT-oss-120b.
\addcontentsline{toc}{subsubsection}{Finding 1: Language models remain weak at this task.}

\textbf{\textcolor{sbase03}{[ Many models are good at either helping or staying silent, but not both. ]}} This is reflected in the large gaps between Demand and No-Dem.\ performance: for example, Qwen3.5-Flash scores 29.1 vs.\ 93.1, Claude-Haiku-4.5 38.9 vs.\ 93.4, and DeepSeek-V3.2 32.3 vs.\ 90.9. These models are \textbf{cautious, but poorly calibrated when intervention is actually needed}. As a result, high No-Dem.\ accuracy does not translate into high overall utility. Even GPT-5-Mini shows a sizable imbalance (66.5 vs.\ 88.0), which limits its average score. Overall, the results show that \textbf{useful assistance requires balanced calibration}: models must \textbf{intervene when needed} and \textbf{stay silent when not}.
\addcontentsline{toc}{subsubsection}{Finding 2: Many models are good at either helping or staying silent, but not both.}

\textbf{\textcolor{sbase03}{[ Targeted training substantially improves this capability. ]}} IntentFlow achieves the best Avg.\ score in the table, 84.2, outperforming Gemini-3-Flash by 3.4 points, GPT-5-Mini by 7.0, and GPT-5-Nano by 12.7. More importantly, this gain does not come from over-predicting assistance: IntentFlow remains strong on both Demand (83.1) and No-Dem.\ (85.2), making it \textbf{the most balanced model overall}. On Demand, it surpasses GPT-5-Mini across all domains, with especially large gains in Program Tutorial, Personal Life, and Content Knowledge, and reaches the best score in several categories. At the same time, it remains slightly below Gemini-3-Flash on overall Demand (83.1 vs.\ 83.3), suggesting that \textbf{even with task-specific training, the hardest intervention cases are not yet fully solved}.
\addcontentsline{toc}{subsubsection}{Finding 3: Targeted training substantially improves this capability.}

\input{pasktables/demand_type}
\subsection{Is LLM A Better Assistant or A Tutor?}
We next examine model performance across two demand types: \textbf{required-type demands (Req.)} and \textbf{insight-type demands (Ins.)}. Req.\ covers \textbf{explicit, goal-directed needs}, such as keyword explanation or factual research, while Ins.\ reflects \textbf{more suggestive and cognitively supportive interactions}, where the user seeks interpretation, guidance, or help grounded in additional context. Together, these two demand types test whether an LLM can function not only as a \textbf{good friend}, but also as a \textbf{mentor}. The results are reported in Table~\ref{tab:demand_type}. We summarize three findings.

\input{pasktables/compare_result}

\textbf{\textcolor{sbase03}{[\space Proprietary frontier models remain stronger in high-value work and learning scenarios. ]}}
In Table~\ref{tab:demand_type}, many of the best results in Work and Learning are achieved by Gemini-3-Flash and the GPT-5 family; for example, Gemini-3-Flash attains the highest average scores in both Work (91.5) and Learning (89.5). This suggests that in domains with clearer utility and denser knowledge requirements, frontier closed models still hold an advantage.
\addcontentsline{toc}{subsubsection}{Finding 1: Proprietary frontier models remain stronger in high-value work and learning scenarios.}

\textbf{\textcolor{sbase03}{[ Req.\ and Ins.\ appear similarly difficult overall, but the gap varies across models and domains. ]}} Although the overall differences between the two demand types are usually small (e.g., 64.1 vs.\ 66.5 for GPT-5-Mini, 82.9 vs.\ 83.3 for Gemini-3-Flash, and 84.0 vs.\ 83.4 for IntentFlow), the relative pattern is much less stable at the domain level. This indicates that the Req./Ins.\ distinction does not lead to a consistent ranking change, but instead interacts with model-specific strengths and application context.
\addcontentsline{toc}{subsubsection}{Finding 2: Req.\ and Ins.\ appear similarly difficult overall, but the gap varies across models and domains.}

\textbf{\textcolor{sbase03}{[ IntentFlow appears more competitive in daily scenarios than in work settings. ]}} It achieves a higher average score in Daily than Gemini-3-Flash (82.3 vs.\ 74.9), but remains behind in Work (84.8 vs.\ 91.5). Overall, IntentFlow slightly surpasses Gemini-3-Flash in average performance (83.7 vs.\ 83.1), suggesting a comparatively balanced profile across everyday user demands.
\addcontentsline{toc}{subsubsection}{Finding 3: IntentFlow appears more competitive in daily scenarios than in work settings.}

\subsection{Multi-round Analysis}
We next study model behavior in realistic multi-turn interactions up to 60 turns (approximately 30 minutes). Detailed results are reported in Table~\ref{tab:turn_degradation}, and the corresponding trends are shown in Figure~\ref{fig:multi_turn_combined}. We summarize three main findings.

\textbf{\textcolor{sbase03}{[ Huge models often exhibit a warm-up effect in early turns. ]}}
Rather than degrading immediately, some frontier models improve after the first bucket: for example, Gemini-3-Flash rises from 85.6 at turns 1--4 to 89.2 at turns 9--12. This suggests that strong models may benefit from an early adaptation phase, during which they accumulate context and produce more targeted responses.
\addcontentsline{toc}{subsubsection}{Finding 1: Large models often exhibit a warm-up effect in early turns.}

\input{pasktables/latency_table}

\textbf{\textcolor{sbase03}{[ Smaller models show clearer degradation as interactions become longer. ]}}
Gemini-2.5-Flash-Lite shows the sharpest decline, with its balanced average dropping from 56.2 to 48.7 and its demand-turn accuracy falling from 30.3 to 7.8 (\(\Delta=-74.1\%\)); similar but milder declines are observed for Claude-Haiku-4.5 (63.1 to 60.7) and GPT-5-Mini (86.7 to 70.3, \(\Delta=-19.0\%\)). These results suggest that smaller models are more vulnerable to long-horizon context accumulation and interaction drift.
\addcontentsline{toc}{subsubsection}{Finding 2: Smaller models show clearer degradation as interactions become longer.}

\textbf{\textcolor{sbase03}{[ IntentFlow maintains stable performance with Pask-MM. ]}}
Although IntentFlow does not start with the highest score in the first bucket (86.1, versus 86.7 for GPT-5-Mini), it maintains a relatively stable performance trajectory with Pask-MM, declining by only 5.0\% from 86.1 to 81.8, compared with -19.0\% for GPT-5-Mini and -17.3\% for Gemini-3-Flash. It also remains competitive in later stages, reaching 83.8 at turns 33--36, 83.4 at turns 37--40, and 81.8 at turns 57--60. These results suggest that Pask-MM may help preserve a more stable workspace over extended interactions.
\addcontentsline{toc}{subsubsection}{Finding 3: IntentFlow maintains stable performance with Pask-MM over long interactions.}

\newpage
\subsection{Additional Analyses}

\begin{figure*}[t]
\centering
\includegraphics[width=\linewidth]{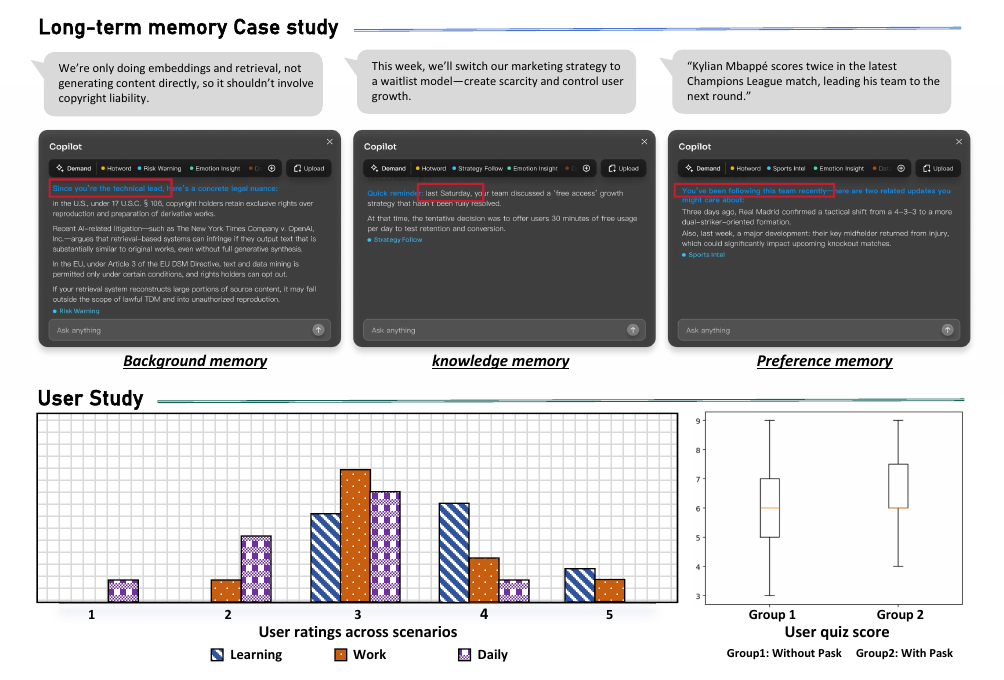}
\caption{\textbf{Long-term memory case study and user study.} \emph{Top:} Case studies illustrating three types of long-term memory in \textsc{Pask-MM}: (1) background memory for understanding user roles and needs, (2) knowledge memory via $\mathcal{M}_{\text{global}}$ for retrieving past facts, and (3) preference memory for personalization. \emph{Bottom-left:} User ratings (20 participants per scenario) are mostly around 3--4, with \textbf{learning} rated highest, \textbf{work} moderate, and \textbf{daily} lowest with some scores near 1--2, highlighting increasing difficulty from structured to open-ended settings. \emph{Bottom-right:} Quiz results (5 participants per group) show that \textsc{Pask} improves average scores from about 6 to 7--7.5 after a 5-minute learning task, demonstrating its effectiveness in knowledge acquisition.}
\label{fig:final-userstudy}
\end{figure*}

\subsubsection{Latency Analysis}
Across models, average per-turn latency typically falls within a few seconds. For instance, GPT-5-Mini averages about 7.1--8.4\,s and GPT-5-Nano about 6.0--6.9\,s, while Gemini-3-Flash and Claude-Haiku-4.5 are faster at around 3.2--4.4\,s and 3.1--3.7\,s, respectively. Gemini-2.5-Flash-Lite further reduces latency to about 1.9--2.4\,s. In contrast, some models exhibit substantially higher latency, such as GPT-oss-120b (7.3--8.7\,s) and Qwen3.5-Flash (17.2--18.6\,s), likely due to additional routing or orchestration overhead. Notably, IntentFlow is consistently the fastest, with latency around 1.3--1.5\,s, which we attribute to its smaller number of activated parameters and more efficient inference.

\subsubsection{Long-Term Memory Analysis}
Evaluating long-term memory in open-ended interactive settings remains inherently challenging, as its benefits are often qualitative and context-dependent. Therefore, instead of relying solely on quantitative metrics, we present a case study analysis to illustrate the practical value of long-term memory in \textsc{Pask-MM}, as shown in Figure~\ref{fig:final-userstudy}.

We categorize the contributions of long-term memory into three main types:
1) \textbf{User background memory}: \textsc{Pask-MM} can retain information about the user’s role, context, and ongoing needs, enabling it to better anticipate and provide relevant assistance. 
2) \textbf{Knowledge memory}: the model leverages $\mathcal{M}_{\text{global}}$ to retrieve previously observed facts and accumulated knowledge across interactions, improving consistency and continuity over time. 
3) \textbf{Preference memory}: remembering user preferences is critical for personalized proactive AI, allowing the system to adapt its responses and suggestions to better align with individual user habits and expectations.

\subsubsection{User Study}

Figure~\ref{fig:final-userstudy} also presents the user study results. The left bar chart shows ratings from three user groups (20 participants each) across different scenarios. Overall, the average scores are concentrated around 3--4. Specifically, the \textbf{learning} scenario peaks around score 4 with most ratings between 3 and 5, indicating the best performance. The \textbf{work} scenario centers around 3--4 with fewer high scores, while the \textbf{daily} scenario is skewed lower, with many ratings around 2--3 and even some at 1. This suggests that learning tasks are relatively easier to support, work scenarios still require stronger context integration, and daily-life assistance remains the most challenging.

The right plot compares two groups (5 participants each) on a quiz after watching a 5-minute educational video. The group without \textsc{Pask} achieves an average score of around 6, while the group with \textsc{Pask} improves to approximately 6--7.5, showing a clear gain of about +1 to +1.5 points. This might indicates that \textsc{Pask} can enhance knowledge acquisition in learning settings.

%% file: pasktables/main_result.tex
\begin{table*}[t]
\centering
\setlength{\tabcolsep}{4pt}
\resizebox{\linewidth}{!}{%
\begin{tabular}{@{}l|l|cccc|ccc|ccc|c@{}}
\toprule
\multirow{3}{*}{\textbf{Model}} & \multirow{3}{*}{\textbf{Type}} &
\multicolumn{4}{c|}{\textbf{Work}} &
\multicolumn{3}{c|}{\textbf{Learning}} &
\multicolumn{3}{c|}{\textbf{Daily}} &
\multirow{3}{*}{\textbf{Overall}} \\
\cmidrule(lr){3-6} \cmidrule(lr){7-9} \cmidrule(lr){10-12}
& & \footnotesize Business & \footnotesize Product & \footnotesize Tech & \footnotesize Work & \footnotesize STEM & \footnotesize Program. & \footnotesize Human. & \footnotesize Personal & \footnotesize Tools \& & \footnotesize Content & \\
& & \footnotesize Metrics & \footnotesize Strategy & \footnotesize Engineer. & \footnotesize Collab. & \footnotesize Lecture & \footnotesize Tutorial & \footnotesize Business & \footnotesize Life & \footnotesize Workflow & \footnotesize Knowl. & \\
\midrule

\multirow{3}{*}{GPT-5-Mini}
& Demand.    & 71.1   & 69.1   & 73.2   & 65.8   & 78.3   & 60.3   & 69.5   & 63.4   & 67.3   & 46.5   & 66.5 \\
& No-Dem.    & 88.8   & 92.5   & 83.7   & 76.8   & 79.2   & 91.2   & 91.7   & 91.2   & 93.8   & 90.6   & 88.0 \\
& \grayrow{Avg.} & \grayrow{\textit{79.9}}   & \grayrow{80.8}   & \grayrow{78.5}   & \grayrow{71.3}   & \grayrow{78.8}   & \grayrow{75.8}   & \grayrow{80.6}   & \grayrow{77.3}   & \grayrow{80.5}   & \grayrow{68.5}   & \grayrow{77.2} \\
\midrule

\multirow{3}{*}{GPT-5-Nano}
& Demand.    & \textit{80.5}   & \textit{87.4}   & \textbf{88.4}   & 77.2   & 58.0   & 53.4   & 51.6   & \textit{77.5}   & 74.4   & 63.1   & 71.2 \\
& No-Dem.    & 52.7   & 70.3   & 65.7   & 50.5   & 82.5   & 89.4   & 87.7   & 74.7   & 75.2   & 69.6   & 71.8 \\
& \grayrow{Avg.} & \grayrow{66.6}   & \grayrow{78.8}   & \grayrow{77.1}   & \grayrow{63.9}   & \grayrow{70.2}   & \grayrow{71.4}   & \grayrow{69.7}   & \grayrow{76.1}   & \grayrow{74.8}   & \grayrow{66.3}   & \grayrow{71.5} \\
\midrule

\multirow{3}{*}{GPT-oss-120b}
& Demand.    & 68.6   & 62.2   & 61.1   & 58.0   & 56.6   & 66.4   & 50.5   & 49.3   & 52.4   & 45.9   & 57.1 \\
& No-Dem.    & 81.9   & 89.5   & 77.3   & 75.0   & 80.9   & 78.8   & 86.7   & 88.5   & 86.2   & 89.6   & 83.4 \\
& \grayrow{Avg.} & \grayrow{75.2}   & \grayrow{75.8}   & \grayrow{69.2}   & \grayrow{66.5}   & \grayrow{68.8}   & \grayrow{72.6}   & \grayrow{68.6}   & \grayrow{68.9}   & \grayrow{69.3}   & \grayrow{67.8}   & \grayrow{70.3} \\
\midrule

\multirow{3}{*}{Gemini-3-Flash}
& Demand.    & \textbf{85.5}   & \textbf{90.3}   & \textit{86.4}   & \textit{78.8}   & \textit{83.2}   & \textbf{89.7}   & \textbf{88.4}   & 76.1   & \textbf{89.9}   & \textit{65.0}   & \textbf{83.3} \\
& No-Dem.    & 63.9   & 79.9   & 84.9   & 67.3   & 84.7   & 86.2   & 81.3   & 79.7   & 80.5   & 74.8   & 78.3 \\
& \grayrow{Avg.} & \grayrow{74.7}   & \grayrow{\textit{85.1}}   & \grayrow{\textit{85.7}}   & \grayrow{\textit{73.0}}   & \grayrow{\textit{84.0}}   & \grayrow{\textit{88.0}}   & \grayrow{\textit{84.8}}   & \grayrow{\textit{77.9}}   & \grayrow{\textit{85.2}}   & \grayrow{\textit{69.9}}   & \grayrow{\textit{80.8}} \\
\midrule

\multirow{3}{*}{Gemini-2.5-Flash-Lite}
& Demand.    & 17.6   & 30.2   & 15.2   & 17.1   & 15.4   & 18.1   & 11.6   & 27.5   & 28.6   & 6.4   & 18.8 \\
& No-Dem.    & 81.2   & 80.8   & 80.8   & 85.0   & 90.2   & 87.5   & 92.3   & 90.8   & 81.9   & 83.8   & 85.4 \\
& \grayrow{Avg.} & \grayrow{49.4}   & \grayrow{55.5}   & \grayrow{48.0}   & \grayrow{51.0}   & \grayrow{52.8}   & \grayrow{52.8}   & \grayrow{51.9}   & \grayrow{59.1}   & \grayrow{55.2}   & \grayrow{45.1}   & \grayrow{52.1} \\
\midrule

\multirow{3}{*}{Claude-Haiku-4.5}
& Demand.    & 62.9   & 57.2   & 55.6   & 45.1   & 21.7   & 39.7   & 15.8   & 32.4   & 36.9   & 21.7   & 38.9 \\
& No-Dem.    & 78.0   & \textit{93.7}   & 90.1   & \textit{88.2}   & 95.6   & \textit{96.9}   & \textbf{99.3}   & \textbf{95.4}   & \textbf{99.0}   & \textbf{98.1}   & \textbf{93.4} \\
& \grayrow{Avg.} & \grayrow{70.5}   & \grayrow{75.5}   & \grayrow{72.8}   & \grayrow{66.7}   & \grayrow{58.6}   & \grayrow{68.3}   & \grayrow{57.5}   & \grayrow{63.9}   & \grayrow{68.0}   & \grayrow{59.9}   & \grayrow{66.2} \\
\midrule

\multirow{3}{*}{DeepSeek-V3.2}
& Demand.    & 44.7   & 42.4   & 33.8   & 34.7   & 7.0   & 18.1   & 12.6   & 53.5   & 54.2   & 21.7   & 32.3 \\
& No-Dem.    & 76.2   & 90.4   & \textit{90.7}   & 85.5   & \textbf{100.0}   & \textbf{96.9}   & \textit{97.3}   & 88.5   & 88.1   & \textit{95.5}   & 90.9 \\
& \grayrow{Avg.} & \grayrow{60.5}   & \grayrow{66.4}   & \grayrow{62.2}   & \grayrow{60.1}   & \grayrow{53.5}   & \grayrow{57.5}   & \grayrow{54.9}   & \grayrow{71.0}   & \grayrow{71.2}   & \grayrow{58.6}   & \grayrow{61.6} \\
\midrule

\multirow{3}{*}{Qwen3.5-Flash}
& Demand.    & 34.6   & 32.0   & 23.2   & 20.2   & 30.8   & 31.9   & 29.5   & 29.6   & 28.6   & 30.6   & 29.1 \\
& No-Dem.    & \textbf{92.8}   & \textbf{94.1}   & \textbf{93.0}   & \textbf{92.7}   & 91.3   & 95.0   & 96.3   & 92.2   & \textit{94.3}   & 89.0   & \textit{93.1} \\
& \grayrow{Avg.} & \grayrow{63.7}   & \grayrow{63.0}   & \grayrow{58.1}   & \grayrow{56.5}   & \grayrow{61.0}   & \grayrow{63.5}   & \grayrow{62.9}   & \grayrow{60.9}   & \grayrow{61.5}   & \grayrow{59.8}   & \grayrow{61.1} \\
\midrule

\multirow{3}{*}{Qwen3-30B-A3B}
& Demand.    & 15.1   & 29.1   & 51.5   & 27.5   & 19.6   & 43.1   & 8.4   & 24.6   & 53.6   & 15.9   & 28.8 \\
& No-Dem.    & \textit{89.5}   & 92.5   & 77.9   & 86.4   & \textit{96.2}   & 78.1   & 97.0   & \textit{94.5}   & 87.6   & 89.0   & 88.9 \\
& \grayrow{Avg.} & \grayrow{52.3}   & \grayrow{60.8}   & \grayrow{64.7}   & \grayrow{57.0}   & \grayrow{57.9}   & \grayrow{60.6}   & \grayrow{52.7}   & \grayrow{59.5}   & \grayrow{70.6}   & \grayrow{52.5}   & \grayrow{58.9} \\
\midrule

\multirow{3}{*}{\textbf{IntentFlow}}
& Demand.    & 79.2   & 86.0   & 85.4   & \textbf{78.8}   & \textbf{85.3}   & \textit{86.2}   & \textit{86.3}   & \textbf{82.4}   & \textit{85.7}   & \textbf{75.8}   & \textit{83.1} \\
& No-Dem.    & 81.9   & 88.3   & 88.4   & 80.0   & 86.9   & 90.0   & 86.7   & 84.8   & 87.1   & 78.3   & 85.2 \\
& \grayrow{Avg.} & \grayrow{\textbf{80.6}}   & \grayrow{\textbf{87.2}}   & \grayrow{\textbf{86.9}}   & \grayrow{\textbf{79.4}}   & \grayrow{\textbf{86.1}}   & \grayrow{\textbf{88.1}}   & \grayrow{\textbf{86.5}}   & \grayrow{\textbf{83.6}}   & \grayrow{\textbf{86.4}}   & \grayrow{\textbf{77.0}}   & \grayrow{\textbf{84.2}} \\

\bottomrule
\end{tabular}%
}
\caption{Main results on the \textsc{IntentFlow} Proactive Demand Detection Benchmark.
Each cell reports the turn-level accuracy score (0--100).
\textit{Demand} = accuracy on demand turns;
\textit{No-Dem.} = accuracy on non-demand turns;
\textit{Avg.} = balanced average (1:1 demand/non-demand).
Columns are grouped by domain:
\textbf{Work} (Business Metrics, Product Strategy, Tech Engineer., Work Collab.),
\textbf{Learning} (STEM Lecture, Program.\ Tutorial, Human.\ Business),
\textbf{Daily} (Personal Life, Tools \& Workflow, Content \& Knowl.).
\textbf{Bold} = best per column; \textit{Italic} = second best.}
\label{tab:main_results}
\end{table*}

%% file: pasktables/demand_type.tex
\begin{table*}[t]
\centering
\setlength{\tabcolsep}{4pt}
\resizebox{\linewidth}{!}{%
\begin{tabular}{@{}l|l|cccc|ccc|ccc|c@{}}
\toprule
\multirow{3}{*}{\textbf{Model}} & \multirow{3}{*}{\textbf{Demand}} &
\multicolumn{4}{c|}{\textbf{Work}} &
\multicolumn{3}{c|}{\textbf{Learning}} &
\multicolumn{3}{c|}{\textbf{Daily}} &
\multirow{3}{*}{\textbf{Overall}} \\
\cmidrule(lr){3-6} \cmidrule(lr){7-9} \cmidrule(lr){10-12}
& & \footnotesize Business & \footnotesize Product & \footnotesize Tech & \footnotesize Work & \footnotesize STEM & \footnotesize Program. & \footnotesize Human. & \footnotesize Personal & \footnotesize Tools \& & \footnotesize Content & \\
& & \footnotesize Metrics & \footnotesize Strategy & \footnotesize Engineer. & \footnotesize Collab. & \footnotesize Lecture & \footnotesize Tutorial & \footnotesize Business & \footnotesize Life & \footnotesize Workflow & \footnotesize Knowl. & \\
\midrule

\multirow{3}{*}{GPT-5-Mini}
& Req.        & 72.9   & 71.6   & 71.0   & 66.9   & \textbf{100.0}   & 49.1   & 33.3   & 60.7   & 67.8   & 47.8   & 64.1 \\
& Ins.        & 65.9   & 60.3   & 83.3   & 63.5   & 75.6   & 70.5   & 70.7   & 65.1   & 64.0   & 46.4   & 66.5 \\
& \grayrow{Avg.} & \grayrow{69.4}   & \grayrow{65.9}   & \grayrow{77.2}   & \grayrow{65.2}   & \grayrow{\textbf{87.8}}   & \grayrow{59.8}   & \grayrow{52.0}   & \grayrow{62.9}   & \grayrow{65.9}   & \grayrow{47.1}   & \grayrow{65.3} \\
\midrule

\multirow{3}{*}{GPT-5-Nano}
& Req.        & \textit{80.5}   & 87.0   & \textbf{88.9}   & \textit{76.2}   & 68.8   & 50.9   & 33.3   & \textit{80.4}   & 77.6   & 60.9   & 70.5 \\
& Ins.        & \textit{80.5}   & \textit{88.9}   & \textit{86.1}   & 79.4   & 56.7   & 55.7   & 52.2   & 75.6   & 56.0   & 63.6   & 69.5 \\
& \grayrow{Avg.} & \grayrow{80.5}   & \grayrow{\textit{88.0}}   & \grayrow{\textbf{87.5}}   & \grayrow{77.8}   & \grayrow{62.8}   & \grayrow{53.3}   & \grayrow{42.8}   & \grayrow{\textit{78.0}}   & \grayrow{66.8}   & \grayrow{62.2}   & \grayrow{70.0} \\
\midrule

\multirow{3}{*}{GPT-oss-120b}
& Req.        & 66.9   & 63.3   & 59.9   & 61.5   & 50.0   & 60.0   & 66.7   & 50.0   & 47.6   & 47.8   & 57.4 \\
& Ins.        & 73.2   & 58.7   & 66.7   & 50.8   & 57.5   & 72.1   & 50.0   & 48.8   & 80.0   & 45.5   & 60.3 \\
& \grayrow{Avg.} & \grayrow{70.1}   & \grayrow{61.0}   & \grayrow{63.3}   & \grayrow{56.1}   & \grayrow{53.8}   & \grayrow{66.0}   & \grayrow{58.4}   & \grayrow{49.4}   & \grayrow{63.8}   & \grayrow{46.6}   & \grayrow{58.9} \\
\midrule

\multirow{3}{*}{Gemini-3-Flash}
& Req.        & \textbf{88.1}   & \textbf{89.3}   & \textit{87.7}   & 76.2   & 75.0   & \textbf{87.3}   & 100.0   & 69.6   & \textbf{90.2}   & \textit{65.2}   & \textit{82.9} \\
& Ins.        & 78.0   & \textbf{93.7}   & 80.6   & \textbf{84.1}   & \textit{84.3}   & \textbf{91.8}   & \textbf{88.0}   & \textit{80.2}   & \textbf{88.0}   & \textit{64.5}   & \textit{83.3} \\
& \grayrow{Avg.} & \grayrow{\textbf{83.0}}   & \grayrow{\textbf{91.5}}   & \grayrow{84.2}   & \grayrow{\textbf{80.2}}   & \grayrow{79.7}   & \grayrow{\textbf{89.5}}   & \grayrow{\textbf{94.0}}   & \grayrow{74.9}   & \grayrow{\textbf{89.1}}   & \grayrow{\textit{64.8}}   & \grayrow{\textit{83.1}} \\
\midrule

\multirow{3}{*}{Gemini-2.5-Flash-Lite}
& Req.        & 14.4   & 30.2   & 16.7   & 15.4   & 0.0   & 12.7   & 0.0   & 28.6   & 25.9   & 0.0   & 14.4 \\
& Ins.        & 26.8   & 30.2   & 8.3   & 20.6   & 17.3   & 23.0   & 12.0   & 26.7   & 44.0   & 9.1   & 21.8 \\
& \grayrow{Avg.} & \grayrow{20.6}   & \grayrow{30.2}   & \grayrow{12.5}   & \grayrow{18.0}   & \grayrow{8.7}   & \grayrow{17.9}   & \grayrow{6.0}   & \grayrow{27.6}   & \grayrow{35.0}   & \grayrow{4.5}   & \grayrow{18.1} \\
\midrule

\multirow{3}{*}{Claude-Haiku-4.5}
& Req.        & 57.6   & 55.3   & 56.8   & 42.3   & 12.5   & 30.9   & 0.0   & 26.8   & 40.6   & 13.0   & 33.6 \\
& Ins.        & 78.0   & 63.5   & 50.0   & 50.8   & 22.8   & 47.5   & 16.3   & 36.0   & 16.0   & 25.5   & 40.6 \\
& \grayrow{Avg.} & \grayrow{67.8}   & \grayrow{59.4}   & \grayrow{53.4}   & \grayrow{46.5}   & \grayrow{17.6}   & \grayrow{39.2}   & \grayrow{8.2}   & \grayrow{31.4}   & \grayrow{28.3}   & \grayrow{19.2}   & \grayrow{37.1} \\
\midrule

\multirow{3}{*}{DeepSeek-V3.2}
& Req.        & 46.6   & 43.3   & 37.0   & 39.2   & 12.5   & 5.5   & 0.0   & 62.5   & 51.7   & 17.4   & 31.6 \\
& Ins.        & 39.0   & 39.7   & 19.4   & 25.4   & 6.3   & 29.5   & 13.0   & 47.7   & 68.0   & 23.6   & 31.2 \\
& \grayrow{Avg.} & \grayrow{42.8}   & \grayrow{41.5}   & \grayrow{28.2}   & \grayrow{32.3}   & \grayrow{9.4}   & \grayrow{17.5}   & \grayrow{6.5}   & \grayrow{55.1}   & \grayrow{59.9}   & \grayrow{20.5}   & \grayrow{31.4} \\
\midrule

\multirow{3}{*}{Qwen3.5-Flash}
& Req.        & 33.1   & 34.4   & 23.5   & 19.2   & 43.8   & 32.7   & \textbf{100.0}   & 25.0   & 28.7   & 23.9   & 36.4 \\
& Ins.        & 39.0   & 23.8   & 22.2   & 22.2   & 29.1   & 31.1   & 27.2   & 32.6   & 28.0   & 33.6   & 28.9 \\
& \grayrow{Avg.} & \grayrow{36.0}   & \grayrow{29.1}   & \grayrow{22.9}   & \grayrow{20.7}   & \grayrow{36.5}   & \grayrow{31.9}   & \grayrow{63.6}   & \grayrow{28.8}   & \grayrow{28.4}   & \grayrow{28.8}   & \grayrow{32.7} \\
\midrule

\multirow{3}{*}{Qwen3-30B-A3B}
& Req.        & 16.9   & 29.8   & 51.9   & 24.6   & 6.2   & 40.0   & 0.0   & 30.4   & 54.5   & 26.1   & 28.0 \\
& Ins.        & 9.8   & 27.0   & 50.0   & 33.3   & 21.3   & 45.9   & 8.7   & 20.9   & 48.0   & 11.8   & 27.7 \\
& \grayrow{Avg.} & \grayrow{13.3}   & \grayrow{28.4}   & \grayrow{51.0}   & \grayrow{28.9}   & \grayrow{13.8}   & \grayrow{43.0}   & \grayrow{4.3}   & \grayrow{25.6}   & \grayrow{51.2}   & \grayrow{19.0}   & \grayrow{27.9} \\
\midrule

\multirow{3}{*}{\textbf{IntentFlow}}
& Req.        & 78.0   & \textit{87.0}   & 85.2   & \textbf{76.9}   & \textit{81.2}   & \textit{85.5}   & \textit{100.0}   & \textbf{82.1}   & \textit{86.0}   & \textbf{78.3}   & \textbf{84.0} \\
& Ins.        & \textbf{82.9}   & 82.5   & \textbf{86.1}   & \textit{82.5}   & \textbf{85.8}   & \textit{86.9}   & \textit{85.9}   & \textbf{82.6}   & \textit{84.0}   & \textbf{74.5}   & \textbf{83.4} \\
& \grayrow{Avg.} & \grayrow{\textit{80.5}}   & \grayrow{84.8}   & \grayrow{\textit{85.7}}   & \grayrow{\textit{79.7}}   & \grayrow{\textit{83.5}}   & \grayrow{\textit{86.2}}   & \grayrow{\textit{93.0}}   & \grayrow{\textbf{82.3}}   & \grayrow{\textit{85.0}}   & \grayrow{\textbf{76.4}}   & \grayrow{\textbf{83.7}} \\

\bottomrule
\end{tabular}%
}
\caption{Performance by demand type on the \textsc{IntentFlow} benchmark.
Each cell reports the turn-level accuracy score (0--100) on demand turns only.
\textit{Req.} = Requirement-type demands (decision support, task planning, problem solving, summarization, information lookup);
\textit{Ins.} = Insight-type demands (risk warning, knowledge gap, callback reminder, context synthesis, trend insight, sentiment analysis).
\textbf{Bold} = best per column; \textit{Italic} = second best.}
\label{tab:demand_type}
\end{table*}

%% file: pasktables/compare_result.tex
\begin{table*}[t]
\centering
\setlength{\tabcolsep}{3pt}
\resizebox{\linewidth}{!}{%
\begin{tabular}{@{}l|l|ccccccccccccccc|c@{}}
\toprule
\multirow{2}{*}{\textbf{Model}} & \multirow{2}{*}{\textbf{Type}} & \multicolumn{15}{c|}{\textbf{Turn Position (bucket)}} & \multirow{2}{*}{$\Delta$(\%)} \\
\cmidrule(lr){3-17}
& & \footnotesize 1-4 & \footnotesize 5-8 & \footnotesize 9-12 & \footnotesize 13-16 & \footnotesize 17-20 & \footnotesize 21-24 & \footnotesize 25-28 & \footnotesize 29-32 & \footnotesize 33-36 & \footnotesize 37-40 & \footnotesize 41-44 & \footnotesize 45-48 & \footnotesize 49-52 & \footnotesize 53-56 & \footnotesize 57-60 & \\
\midrule
\multirow{3}{*}{GPT-5-Mini}
& Demand. & 80.3 & 76.6 & 74.7 & 68.4 & 69.3 & 67.2 & 67.7 & 64.5 & 60.2 & 67.0 & 65.4 & 59.4 & 60.0 & 50.9 & 51.0 & -36.5 \\
& No-Dem. & 93.1 & 92.7 & 86.3 & 87.9 & 82.8 & 90.5 & 84.2 & 85.5 & 89.1 & 87.2 & 83.6 & 91.4 & 86.7 & \textit{95.6} & 89.6 & -3.8 \\
& \grayrow{Avg.} & \grayrow{\textbf{86.7}} & \grayrow{84.6} & \grayrow{80.5} & \grayrow{78.1} & \grayrow{76.0} & \grayrow{78.9} & \grayrow{76.0} & \grayrow{\textit{75.0}} & \grayrow{74.7} & \grayrow{\textit{77.1}} & \grayrow{74.5} & \grayrow{75.4} & \grayrow{73.4} & \grayrow{73.2} & \grayrow{70.3} & \grayrow{-19.0} \\
\midrule
\multirow{3}{*}{GPT-5-Nano}
& Demand. & \textbf{88.6} & 78.6 & 82.0 & 80.9 & 67.5 & 69.0 & 66.1 & 69.1 & 71.0 & 75.5 & 66.7 & 76.8 & 70.8 & 75.4 & 60.8 & -31.4 \\
& No-Dem. & 75.2 & 68.8 & 77.2 & 73.1 & 71.1 & 71.6 & 71.1 & 73.1 & 69.6 & 67.5 & 70.9 & 68.8 & 74.7 & 72.1 & 77.1 & \textbf{2.5} \\
& \grayrow{Avg.} & \grayrow{81.9} & \grayrow{73.7} & \grayrow{79.6} & \grayrow{77.0} & \grayrow{69.3} & \grayrow{70.3} & \grayrow{68.6} & \grayrow{71.1} & \grayrow{70.3} & \grayrow{71.5} & \grayrow{68.8} & \grayrow{72.8} & \grayrow{72.7} & \grayrow{\textit{73.7}} & \grayrow{68.9} & \grayrow{-15.8} \\
\midrule
\multirow{3}{*}{GPT-oss-120b}
& Demand. & 56.1 & 54.5 & 64.0 & 63.2 & 57.0 & 68.1 & 53.2 & 49.1 & 57.0 & 46.8 & 50.6 & 58.0 & 56.9 & 61.4 & 54.9 & \textbf{-2.1} \\
& No-Dem. & 90.5 & 85.8 & 89.3 & 86.3 & 87.2 & 82.8 & 82.2 & 80.0 & 79.7 & 76.9 & 83.6 & 86.0 & 85.5 & 83.8 & 77.1 & -14.8 \\
& \grayrow{Avg.} & \grayrow{73.3} & \grayrow{70.2} & \grayrow{76.7} & \grayrow{74.7} & \grayrow{72.1} & \grayrow{75.5} & \grayrow{67.7} & \grayrow{64.5} & \grayrow{68.3} & \grayrow{61.9} & \grayrow{67.1} & \grayrow{72.0} & \grayrow{71.2} & \grayrow{72.6} & \grayrow{66.0} & \grayrow{-9.9} \\
\midrule
\multirow{3}{*}{Gemini-3-Flash}
& Demand. & 75.0 & \textit{85.1} & \textbf{92.7} & \textbf{89.7} & \textbf{87.7} & \textbf{86.2} & \textbf{85.5} & \textit{80.0} & \textbf{84.9} & \textbf{86.2} & \textbf{82.7} & \textbf{84.1} & \textbf{86.2} & \textit{77.2} & \textit{68.6} & -8.5 \\
& No-Dem. & 96.2 & 93.1 & 85.8 & 84.1 & 77.8 & 75.1 & 75.7 & 69.7 & 65.2 & 67.5 & 66.4 & 75.3 & 69.9 & 69.1 & 72.9 & -24.2 \\
& \grayrow{Avg.} & \grayrow{85.6} & \grayrow{\textbf{89.1}} & \grayrow{\textbf{89.2}} & \grayrow{\textbf{86.9}} & \grayrow{\textit{82.7}} & \grayrow{\textit{80.7}} & \grayrow{\textit{80.6}} & \grayrow{74.8} & \grayrow{\textit{75.1}} & \grayrow{76.8} & \grayrow{\textit{74.5}} & \grayrow{\textit{79.7}} & \grayrow{\textit{78.0}} & \grayrow{73.2} & \grayrow{\textit{70.8}} & \grayrow{-17.3} \\
\midrule
\multirow{3}{*}{Gemini-2.5-Flash-Lite}
& Demand. & 30.3 & 30.5 & 28.0 & 21.3 & 21.1 & 19.0 & 15.3 & 17.3 & 8.6 & 22.3 & 17.3 & 17.4 & 15.4 & 8.8 & 7.8 & -74.1 \\
& No-Dem. & 82.1 & 78.0 & 83.8 & 83.0 & 85.6 & 87.6 & 87.5 & 88.3 & 89.9 & 89.7 & \textit{90.0} & 91.4 & 90.4 & 86.8 & 89.6 & 9.2 \\
& \grayrow{Avg.} & \grayrow{56.2} & \grayrow{54.3} & \grayrow{55.9} & \grayrow{52.1} & \grayrow{53.3} & \grayrow{53.3} & \grayrow{51.4} & \grayrow{52.8} & \grayrow{49.2} & \grayrow{56.0} & \grayrow{53.6} & \grayrow{54.4} & \grayrow{52.9} & \grayrow{47.8} & \grayrow{48.7} & \grayrow{-13.3} \\
\midrule
\multirow{3}{*}{Claude-Haiku-4.5}
& Demand. & 27.3 & 42.2 & 44.0 & 41.9 & 43.0 & 48.3 & 41.1 & 45.5 & 39.8 & 37.2 & 48.1 & 42.0 & 44.6 & 42.1 & 25.5 & -6.5 \\
& No-Dem. & \textbf{98.9} & \textbf{97.7} & \textit{96.4} & \textbf{96.7} & \textit{92.8} & \textbf{94.7} & \textbf{94.7} & 91.0 & \textit{92.8} & 86.3 & 89.1 & \textbf{95.7} & \textit{92.8} & 92.6 & \textit{95.8} & \textit{-3.1} \\
& \grayrow{Avg.} & \grayrow{63.1} & \grayrow{70.0} & \grayrow{70.2} & \grayrow{69.3} & \grayrow{67.9} & \grayrow{71.5} & \grayrow{67.9} & \grayrow{68.2} & \grayrow{66.3} & \grayrow{61.8} & \grayrow{68.6} & \grayrow{68.9} & \grayrow{68.7} & \grayrow{67.4} & \grayrow{60.7} & \grayrow{\textit{-3.8}} \\
\midrule
\multirow{3}{*}{DeepSeek-V3.2}
& Demand. & 33.3 & 34.4 & 36.7 & 31.6 & 40.4 & 38.8 & 35.5 & 32.7 & 33.3 & 25.5 & 29.6 & 30.4 & 35.4 & 33.3 & 35.3 & 5.9 \\
& No-Dem. & 92.4 & 90.8 & 92.9 & 92.9 & \textbf{92.8} & \textit{93.5} & 92.1 & \textbf{93.1} & 88.4 & 85.5 & 88.2 & 91.4 & 91.6 & 91.2 & \textbf{95.8} & 3.8 \\
& \grayrow{Avg.} & \grayrow{62.8} & \grayrow{62.6} & \grayrow{64.8} & \grayrow{62.2} & \grayrow{66.6} & \grayrow{66.1} & \grayrow{63.8} & \grayrow{62.9} & \grayrow{60.9} & \grayrow{55.5} & \grayrow{58.9} & \grayrow{60.9} & \grayrow{63.5} & \grayrow{62.3} & \grayrow{65.6} & \grayrow{4.3} \\
\midrule
\multirow{3}{*}{Qwen3.5-Flash}
& Demand. & 20.5 & 28.6 & 24.7 & 26.5 & 33.3 & 25.0 & 28.2 & 22.7 & 24.7 & 28.7 & 39.5 & 29.0 & 30.8 & 31.6 & 33.3 & 63.0 \\
& No-Dem. & \textit{98.5} & \textit{97.2} & \textbf{97.0} & \textit{94.0} & 91.1 & 91.1 & 92.1 & 89.7 & 87.7 & \textbf{91.5} & 87.3 & \textit{91.4} & \textbf{94.0} & \textbf{95.6} & 89.6 & -9.0 \\
& \grayrow{Avg.} & \grayrow{59.5} & \grayrow{62.9} & \grayrow{60.8} & \grayrow{60.2} & \grayrow{62.2} & \grayrow{58.1} & \grayrow{60.2} & \grayrow{56.2} & \grayrow{56.2} & \grayrow{60.1} & \grayrow{63.4} & \grayrow{60.2} & \grayrow{62.4} & \grayrow{63.6} & \grayrow{61.5} & \grayrow{\textbf{3.4}} \\
\midrule
\multirow{3}{*}{Qwen3-30B-A3B}
& Demand. & 41.7 & 43.5 & 39.3 & 35.3 & 27.2 & 31.9 & 27.4 & 27.3 & 25.8 & 23.4 & 30.9 & 21.7 & 24.6 & 19.3 & 17.6 & -57.6 \\
& No-Dem. & 85.1 & 83.5 & 86.3 & 88.5 & 91.7 & 87.6 & \textit{92.8} & \textit{92.4} & \textbf{92.8} & \textit{90.6} & \textbf{97.3} & 91.4 & 88.0 & 88.2 & 89.6 & 5.3 \\
& \grayrow{Avg.} & \grayrow{63.4} & \grayrow{63.5} & \grayrow{62.8} & \grayrow{61.9} & \grayrow{59.4} & \grayrow{59.7} & \grayrow{60.1} & \grayrow{59.8} & \grayrow{59.3} & \grayrow{57.0} & \grayrow{64.1} & \grayrow{56.6} & \grayrow{56.3} & \grayrow{53.8} & \grayrow{53.6} & \grayrow{-15.4} \\
\midrule
\multirow{3}{*}{\textbf{IntentFlow}}
& Demand. & \textit{85.6} & \textbf{85.7} & \textit{84.7} & \textit{83.8} & \textit{84.2} & \textit{81.9} & \textit{82.3} & \textbf{81.8} & \textit{82.8} & \textit{83.0} & \textit{81.5} & \textit{82.6} & \textit{80.0} & \textbf{80.7} & \textbf{82.4} & \textit{-3.8} \\
& No-Dem. & 86.6 & 86.2 & 85.3 & 86.3 & 84.4 & 85.8 & 83.6 & 85.5 & 84.8 & 83.8 & 84.5 & 81.7 & 83.1 & 80.9 & 81.2 & -6.2 \\
& \grayrow{Avg.} & \grayrow{\textit{86.1}} & \grayrow{\textit{86.0}} & \grayrow{\textit{85.0}} & \grayrow{\textit{85.0}} & \grayrow{\textbf{84.3}} & \grayrow{\textbf{83.8}} & \grayrow{\textbf{82.9}} & \grayrow{\textbf{83.7}} & \grayrow{\textbf{83.8}} & \grayrow{\textbf{83.4}} & \grayrow{\textbf{83.0}} & \grayrow{\textbf{82.2}} & \grayrow{\textbf{81.6}} & \grayrow{\textbf{80.8}} & \grayrow{\textbf{81.8}} & \grayrow{-5.0} \\

\bottomrule
\end{tabular}%
}
\caption{Per-turn-bucket performance under the multi-turn setting.
Turn positions are grouped into buckets of 4 consecutive turns.
\textit{Demand} = accuracy on demand turns;
\textit{No-Dem.} = accuracy on non-demand turns;
\textit{Avg.} = balanced average.
$\Delta$(\%) = relative change from first to last bucket.
\textbf{Bold} = best per column; \textit{Italic} = second best.}
\label{tab:turn_degradation}
\end{table*}

%% file: pasktables/latency_table.tex
\begin{table*}[t]
\centering
\setlength{\tabcolsep}{4pt}
\resizebox{0.99\linewidth}{!}{%
\begin{tabular}{@{}l|ccc|ccc|ccc|ccc|ccc@{}}
\toprule
\multirow{2}{*}{\textbf{Model}} & \multicolumn{15}{c}{\textbf{Per-Turn Latency (ms)}} \\
\cmidrule(lr){2-16}
& \multicolumn{3}{c|}{$T$=1--12} & \multicolumn{3}{c|}{$T$=13--24} & \multicolumn{3}{c|}{$T$=25--36} & \multicolumn{3}{c|}{$T$=37--48} & \multicolumn{3}{c}{$T$=49--60} \\
\cmidrule(lr){2-4} \cmidrule(lr){5-7} \cmidrule(lr){8-10} \cmidrule(lr){11-13} \cmidrule(lr){14-16}
& \footnotesize Dem. & \footnotesize N-D. & \footnotesize Avg. & \footnotesize Dem. & \footnotesize N-D. & \footnotesize Avg. & \footnotesize Dem. & \footnotesize N-D. & \footnotesize Avg. & \footnotesize Dem. & \footnotesize N-D. & \footnotesize Avg. & \footnotesize Dem. & \footnotesize N-D. & \footnotesize Avg. \\
\midrule
GPT-5-Mini & 10.4k & 6.7k & 8.1k & 10.2k & 7.1k & 8.4k & 9.9k & 7.3k & 8.4k & 9.3k & 6.4k & 7.6k & 8.6k & 5.8k & 7.1k \\
\midrule
GPT-5-Nano & 7.4k & 6.3k & 6.7k & 6.9k & 5.7k & 6.2k & 7.0k & 6.2k & 6.5k & 7.2k & 6.2k & 6.6k & 6.3k & 5.8k & 6.0k \\
\midrule
GPT-oss-120b & 7.7k & 6.3k & 6.8k & 8.3k & 6.6k & 7.3k & 9.1k & 8.4k & 8.7k & 7.9k & 6.8k & 7.3k & 8.6k & 7.4k & 8.0k \\
\midrule
Gemini-3-Flash & 3.6k & 3.0k & 3.2k & 3.9k & 3.7k & 3.8k & 4.3k & 4.1k & 4.2k & 4.3k & 4.3k & 4.3k & 4.4k & 4.3k & 4.4k \\
\midrule
Gemini-2.5-Flash-Lite & 2.8k & 2.2k & 2.4k & 2.3k & 2.4k & 2.4k & 2.3k & 2.3k & 2.3k & 2.2k & 2.1k & 2.2k & 2.0k & 1.9k & 1.9k \\
\midrule
Claude-Haiku-4.5 & 3.6k & 2.8k & 3.1k & 4.0k & 3.4k & 3.6k & 4.1k & 3.2k & 3.6k & 4.1k & 3.4k & 3.7k & 4.0k & 3.5k & 3.7k \\
\midrule
DeepSeek-V3.2 & 3.4k & 3.1k & 3.2k & 3.8k & 3.1k & 3.4k & 3.7k & 3.1k & 3.4k & 3.5k & 3.2k & 3.3k & 4.0k & 3.1k & 3.5k \\
\midrule
Qwen3.5-Flash & 16.1k & 15.7k & 15.9k & 16.6k & 19.9k & 18.6k & 16.7k & 18.7k & 17.8k & 16.4k & 17.8k & 17.2k & 16.4k & 17.8k & 17.2k \\
\midrule
Qwen3-30B-A3B & \textbf{1.4k} & \textbf{988} & \textbf{1.1k} & \textbf{1.4k} & \textbf{1.1k} & \textbf{1.2k} & \textbf{1.5k} & \textbf{1.2k} & \textbf{1.3k} & \textbf{1.2k} & \textbf{1.1k} & \textbf{1.1k} & \textbf{1.2k} & \textbf{873} & \textbf{1.0k} \\
\midrule
\textbf{IntentFlow} & \textit{1.6k} & \textit{1.2k} & \textit{1.3k} & \textit{1.6k} & \textit{1.2k} & \textit{1.4k} & \textit{1.7k} & \textit{1.2k} & \textit{1.4k} & \textit{1.7k} & \textit{1.3k} & \textit{1.5k} & \textit{1.8k} & \textit{1.3k} & \textit{1.5k} \\

\bottomrule
\end{tabular}%
}
\caption{Per-turn inference latency by conversation position.
Turn positions are grouped into buckets of 12 consecutive turns ($T$).
\textit{Dem.} = average latency on demand turns (model generates a response);
\textit{N-D.} = average latency on non-demand turns (model outputs \texttt{[NO\_DEMAND]});
\textit{Avg.} = overall average.
Demand turns consistently incur higher latency due to longer generated outputs.
\textbf{Bold} = fastest; \textit{Italic} = second fastest.}
\label{tab:demand_latency}
\end{table*}

%% file: pasksections/7-relatedWork.tex
\section{Related Work}

\paragraph{Proactive AI agents}
Proactivity is a longstanding objective in intelligent agent research, where systems are expected to anticipate user needs and act beyond explicit instructions. Early work framed this capability through meta-level control and situation awareness~\citep{myers2007proactive,proagent}. With the emergence of LLM-based systems, proactive behavior has become more tractable. Recent efforts fall into two categories: (1) task-level proactivity, where agents identify missing information and iteratively refine plans through interaction~\citep{parimi2024proactiveaisystem, zhang2024ask}; and (2) system-level proactivity, where agents initiate actions outside direct user prompts, such as in programming assistance, computer operation, and collaborative gameplay~\citep{proactive_program, proactive_computer, proagent_gaming}. More general approaches leverage multimodal context to trigger recommendations or interventions~\citep{myers2007proactive,proagent,yang2025contextagent}, but remain largely confined to controlled settings and struggle with complex, long-horizon real-world tasks.

\paragraph{Agents with memory}
LLM agents are increasingly modeled as interactive systems augmented with external memory, enabling both short-term adaptation within tasks and long-term accumulation across tasks~\citep{li2024improving, wan2025rema, yao2022react, chen2023fireact}. Existing approaches typically treat memory as a unified, evolving state, where memory formation, update, and retrieval constitute the core lifecycle~\citep{zhao2024expel, rasmussen2025zep, li2025hello, wan2025rema}. This paradigm has become a general foundation for memory-augmented agents in both single- and multi-agent settings. Our method follows this framework and is particularly inspired by structured external memory designs such as Mem0, but differs in introducing proactive memory management, where memory is updated and utilized in anticipation of future interactions rather than solely in response to past ones.

\paragraph{Streaming models}
IntentFlow is motivated by two complementary directions in streaming modeling: \emph{streaming understanding} and \emph{instant response}. Streaming understanding, studied primarily in video, enables models to incrementally process long inputs and produce intermediate outputs with reduced latency~\citep{chen2024videollm,wang2024videollm,huang2025online}. Instant response has been central in audio and dialogue systems, where real-time interaction is required~\citep{defossez2024moshi, zhang2025stream, xie2024mini, xie2024mini2}. IntentFlow aligns with streaming dialogue models in interaction design, while incorporating streaming understanding mechanisms to maintain deep contextual reasoning over continuously arriving inputs.

%% file: pasksections/8-Conclusion.tex
\section{Conclusion and Next Steps}
\label{sec:conclusion}


We introduce Pask, a proactive AI agent built to test whether LLMs can operate in real-world environments by predicting user needs in real time and delivering meaningful help beyond reactive response. To support this goal, we propose DD-MM-PAS, a general framework that combines demand detection, long-term evolving memory, and a streaming proactive agent. We realize this framework in real-world settings. For demand detection, we develop IntentFlow, including a streaming architecture, a scalable data generation pipeline, and a two-stage training recipe. For memory, we design a hybrid memory that organizes user information across workspace, user, and global levels, capturing progress, preference, and long-term context. We also show that deep proactive behavior depends not only on stronger modeling, but on tight coordination between demand detection and memory under strict latency constraints.

Our results show that, despite the strength of closed-source models, deep intent understanding remains unsolved. Current models can often handle shallow assistance \underline{\textit{“Do you want me to send a message for you?”}}, but they still struggle with the deeper cases that matter more, such as \underline{\textit{ “the user is misunderstanding the boss’s real}}
\vspace{-2mm}

\underline{\textit{ intent, and the agent should step in now”}}. In this setting, IntentFlow does not claim a new ceiling; rather, it shows that with the right training recipe, an open-source model can reach performance close to closed-source models, and in multi-turn real-world settings, even outperform existing open-source baselines. We finally summarize nine key findings based on our experiments and give additional discussion on the problem from the perspectives of latency, long-term memory, and user study. 

Overall, Pask provides a concrete path toward proactive AI agents that combine real-time response with deep user understanding. At the same time, the current capability is still only a first step. The real goal of proactive AI is not basic reminders or surface-level prediction, but the ability to detect needs users have not fully expressed, uncover what is truly important in context, and take actions that deliver genuine value. We hope the framework and benchmark introduced in this work can push the field away from shallow proactive behavior and toward proactive AI that is deep, useful, and real enough to leave the lab and work in the world.

